\documentclass{article} 
\usepackage{iclr2027_conference,times}

\usepackage{amsmath,amsfonts,bm}

\def\eqref#1{equation~\ref{#1}}

\def\1{\bm{1}}

\def\eps{{\epsilon}}

\def\vx{{\bm{x}}}

\DeclareMathAlphabet{\mathsfit}{\encodingdefault}{\sfdefault}{m}{sl}
\SetMathAlphabet{\mathsfit}{bold}{\encodingdefault}{\sfdefault}{bx}{n}

\def\gH{{\mathcal{H}}}

\def\gO{{\mathcal{O}}}
\def\gP{{\mathcal{P}}}

\newcommand{\Ls}{\mathcal{L}}
\newcommand{\R}{\mathbb{R}}

\DeclareMathOperator*{\argmax}{arg\,max}
\DeclareMathOperator*{\argmin}{arg\,min}

\usepackage{hyperref}
\usepackage{url}
\usepackage{booktabs}
\usepackage{graphicx}
\usepackage{amsmath}
\usepackage{enumitem}
\usepackage{multirow}
\usepackage{subcaption}

\title{Less Is More: Genetic Frame Selection for \\ Efficient Novel View Synthesis}

\author{
Diego E. Farchione\textsuperscript{1},
Ramzi Idoughi\textsuperscript{1},
Alberto Jaspe-Villanueva\textsuperscript{1},
Peter Wonka\textsuperscript{1} \\[0.5em]
\textsuperscript{1}King Abdullah University of Science and Technology (KAUST)
}

\usepackage{cuted}
\usepackage{caption}

\usepackage{microtype}          
\makeatletter
\renewcommand{\section}{\@startsection{section}{1}{\z@}%
  {-0.8ex \@plus -0.5ex \@minus -0.2ex}
  {0.4ex \@plus 0.3ex \@minus 0.2ex}
  {\large\sc\raggedright}}
\renewcommand{\subsection}{\@startsection{subsection}{2}{\z@}%
  {-0.7ex \@plus -0.5ex \@minus -0.2ex}
  {0.15ex \@plus .2ex}
  {\normalsize\sc\raggedright}}
\makeatother

\iclrfinalcopy 
\begin{document}

\maketitle

\lhead{Preprint}

\begin{center}
    \includegraphics[width=0.98\textwidth]
    {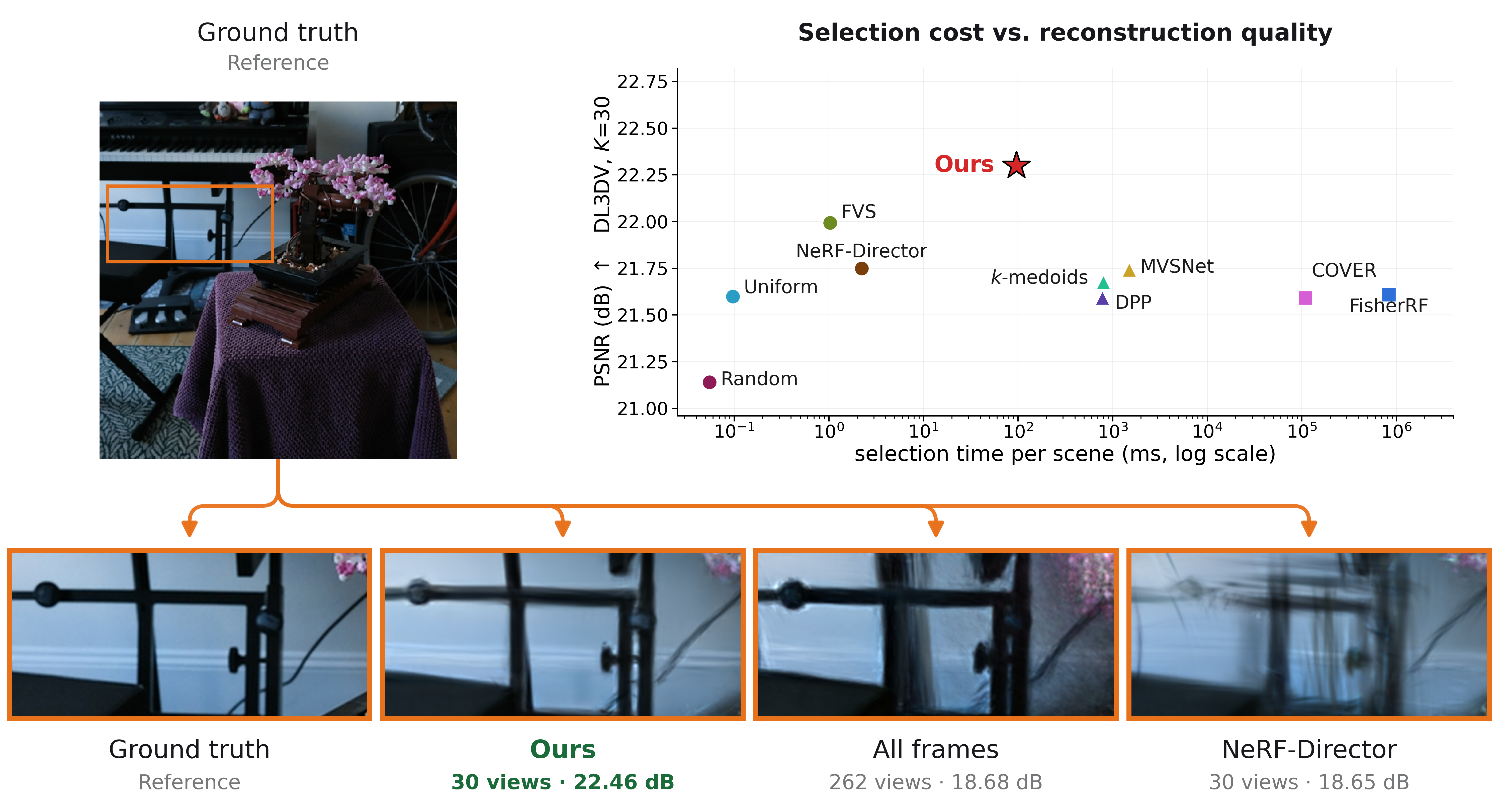}

    \captionof{figure}{\textbf{Overview.} Without appearance features, our method selects the most informative views, 30 in the example shown, in approximately $0.1$\,s, yielding a sharper reconstruction than prior view-selection methods and reconstruction from all available frames.}
    \label{fig:teaser}
\end{center}

\vspace{-0.35em}
\begin{abstract}
Feed-forward novel view synthesis reconstructs a scene from many input images in a single forward pass, yet more views do not necessarily improve performance: redundant or poorly chosen frames increase computational cost and may even degrade reconstruction quality. We address the problem of selecting, from an already captured sequence, a fixed-size subset of input views that is most informative for reconstructing specified target viewpoints.
We propose a render-free view selector that scores candidate frames based on three complementary criteria: target-view coverage, measured against observed frames that stand in for the targets, redundancy with previously selected views, and image sharpness. A lightweight scoring network then selects the most informative frames without rendering, reconstruction, or per-scene optimization at inference time. To train the selector, we distill an expensive offline search procedure in which a genetic algorithm identifies high-quality subsets by directly optimizing reconstruction performance on training scenes. The selector learns to reproduce these choices from geometric and image-level features alone. Across six datasets and multiple input budgets, our method consistently outperforms both geometric and reconstruction-aware view-selection baselines while incurring significantly lower selection costs than reconstruction-based alternatives. Moreover, carefully selected subsets can outperform feed-forward reconstruction from the full input sequence. The learned selector generalizes across diverse reconstruction paradigms (feed-forward, 3D Gaussian Splatting, and NeRF), to object-targeted reconstruction and to a cross-capture setting in which the target views come from a separate acquisition pass. More broadly, our results indicate that explicitly reasoning about target relevance and inter-view redundancy is a fundamental factor in efficient scene reconstruction.
\end{abstract}


\vspace{-0.35em}

\section{Introduction}
\label{sec:intro}

Feed-forward 3D reconstruction models can now recover complete scenes from multi-view images in a single forward pass, reducing what once required hours of per-scene optimization to only a few seconds~\citep{wang2024dust3r,leroy2024mast3r,wang2025vggt,yang2025fast3r, ye2025noposplat,jiang2025anysplat,furutani2026wild3r}. 
As these models scale to a larger set of input images, one might expect reconstruction quality to improve simply by providing more views. In practice, however, using all available frames is neither the most efficient nor the most accurate strategy for current feed-forward reconstructors. Computational cost and memory consumption grow with the number of inputs, while reconstruction quality often saturates or even degrades. Captured videos contain long runs of near-duplicate frames, as well as blurred, poorly posed, or uninformative views that consume model capacity without adding parallax. In our experiments, we find that a carefully chosen subset reconstructs a scene \emph{more} accurately than the full sequence, while using only a fraction of the input.
Selecting such a subset is difficult because a frame has no value on its own: it depends on where the novel views will be rendered, and on which other frames are chosen alongside it. Existing geometric criteria, such as uniform and farthest-view sampling~\citep{xiao2024nerfdirector}, largely ignore these dependencies, relying on camera placement or sparse geometry rather than reconstruction quality. Reconstruction-aware methods based on uncertainty, or information gain~\citep{pan2022activenerf, jiang2024fisherrf, chen2026cover}, are more closely aligned with the reconstruction objective, but they typically evaluate candidates by repeatedly fitting, making them more expensive than the feed-forward pipelines~\citep{ziwen2025long,yang2025fast3r,jiang2025anysplat}.

\begin{figure}[t]
\centering
\includegraphics[width=0.85\textwidth]{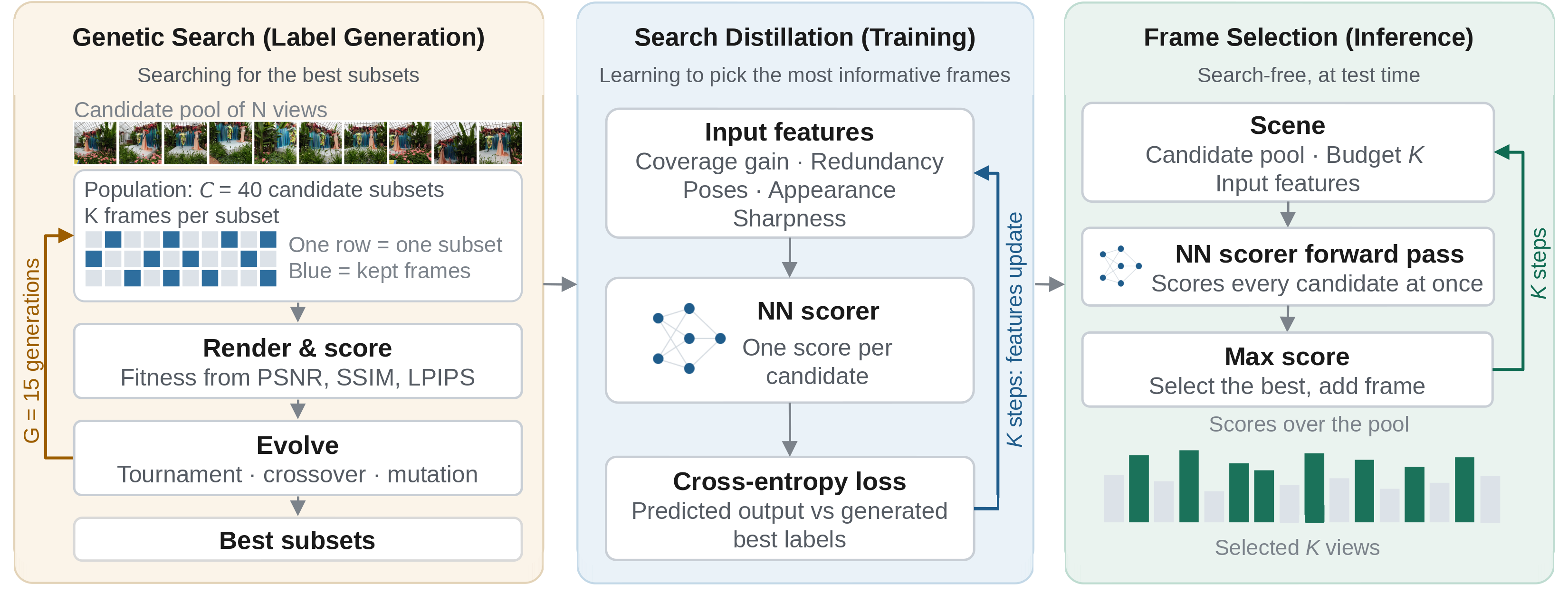}
\caption{\textbf{Overview.} A genetic search over frame subsets, scored by rendered quality, produces best
subsets for each training scene; a compact scorer is distilled from those choices using
coverage, redundancy, pose-to-target and appearance features; at test time it ranks every candidate in one
forward pass and appends one frame at each step. The selector can run in less than one second with no search, no rendering, no per-scene optimization.}
\label{fig:pipeline}
\end{figure}

We propose a \textbf{render-free frame selector} (Figure~\ref{fig:teaser}) that addresses both dependencies while remaining inexpensive at inference time. Given a pool of candidate frames with their camera poses, and the poses at which novel views are to be rendered, our method identifies the observed frames geometrically closest to the targets and uses them as proxies for the target views. A compact network then scores candidate views using features that capture coverage, redundancy, appearance similarity, and pose relationships to the target region. Frames are picked greedily (see Figure~\ref{fig:pipeline}), with selection-conditioned features recomputed after each step, allowing each candidate to be evaluated based on the information it contributes beyond the views already chosen. This procedure requires no rendering, no reconstruction fitting, and no per-scene optimization, being completed in seconds per scene. 
In particular, training supervision comes from the reconstructor itself. Since evaluating a subset requires running a reconstruction, we incur this cost offline: on training scenes, a reconstruction-aware genetic search explores candidate subsets and scores them by rendered quality, producing high-quality reference selections. The selector is then trained through behavior cloning to imitate these choices one greedy step at a time, yielding reconstruction-aware view selection at the cost of a single forward pass per step.

Experiments across outdoor captures, unbounded scenes, and handheld indoor videos show that the selector outperforms both geometric and reconstruction-aware baselines while remaining inexpensive. The same frozen selector, trained on labels from one feed-forward model, also improves others without retraining, suggesting it learns general properties of informative input subsets rather than model-specific behavior, and it extends to object-targeted selection, prioritizing views relevant to a designated object in a cluttered scene.

\textbf{Contributions.}
Feed-forward reconstruction is typically evaluated using a predetermined set of input views, leaving the question of which views to choose largely unexplored. We address this question directly and contribute the following:
\begin{itemize}[leftmargin=1.4em,itemsep=1pt,topsep=2pt]
    \item We introduce a \textbf{reconstruction-aware genetic search} that directly optimizes rendered reconstruction quality over candidate frame subsets. 
  \item We propose a \textbf{render-free frame selector} with \textbf{selection-conditioned features} that accounts for both target relevance and inter-frame redundancy while requiring no rendering, reconstruction fitting, or per-scene optimization at inference time.
  \item We demonstrate that carefully selected subsets can \textbf{outperform the full input sequence} in reconstruction quality for current feed-forward reconstructors while substantially \textbf{reducing computational cost and memory usage}.
  \item We benchmark against a broad range of geometric and reconstruction-aware view-selection strategies (spread-based, co-visibility, triangulation-angle, and information-gain), and demonstrate \textbf{zero-shot transfer across reconstructors} and applicability to \textbf{object-targeted reconstruction}.
\end{itemize}

\vspace{-0.35em}

\section{Related Work}
\label{sec:related}
\textbf{Neural scene reconstruction and novel view synthesis.}
NeRF~\citep{mildenhall2020nerf} demonstrated high-quality novel view synthesis by fitting a radiance field to posed images and rendering through volumetric ray marching. Subsequent work improved training and rendering efficiency through multiresolution hash encodings~\citep{mueller2022instantngp}, extensions to large unbounded scenes~\citep{barron2022mipnerf360}, and explicit scene representations such as 3D Gaussian Splatting~\citep{kerbl20233dgs}. Despite their differences, these methods remain \emph{per-scene} approaches, requiring a dedicated optimization procedure for each capture. Reconstruction quality therefore depends on the provided views, yet their selection is rarely formulated as a reconstruction-aware optimization problem. Throughout, camera poses are assumed to be known, typically from structure-from-motion~\citep{schoenberger2016sfm}.
Recent work replaces per-scene optimization with direct feed-forward prediction. Early approaches such as PixelNeRF~\citep{yu2021pixelnerf} conditioned radiance fields on image features, while later methods directly regress 3D scene representations from multiple views. Examples include GS-LRM~\citep{zhang2024gslrm}, LongLRM~\citep{ziwen2025long}, PixelSplat~\citep{charatan2024pixelsplat}, MVSplat~\citep{chen2024mvsplat} and DepthSplat~\citep{xu2025depthsplat}.
A parallel class of feed-forward reconstructors operates directly on unposed images, jointly recovering camera parameters and scene geometry or predicting Gaussian scene representations~\citep{wang2024dust3r,leroy2024mast3r,wang2025vggt,yang2025fast3r,smart2024splatt3r,ye2025noposplat,jiang2025anysplat,furutani2026wild3r}. We use a model from this family to evaluate whether our selector transfers across reconstructors.
As memory and compute scale with the number of input views, the reconstruction quality often saturates due to frame redundancy, motivating selecting informative subsets of candidate views.

\textbf{View selection.} A large body of work studies view selection through geometric criteria such as camera coverage, visibility, baseline, and triangulation quality~\citep{goesele2007multi, hornung2008image, furukawa2010towards, schoenberger2016mvs, yao2018mvsnet, xiao2024nerfdirector}. These approaches operate primarily on camera placement and sparse geometry, without directly accounting for the reconstruction objective. Reconstruction-aware criteria instead score candidates by uncertainty, information gain, or estimated reconstruction quality~\citep{pan2022activenerf, jiang2024fisherrf, chen2026cover, wilson2025pop, goli2024bayesrays, sunderhauf2023density, li2025ougs, tosi2025warprf}, and are often developed in the active setting~\citep{connolly1985determination, isler2016information}, where a candidate is evaluated before a new observation is acquired and the cost of fitting an intermediate model is therefore justified. A structured taxonomy of next-best-view strategies is given by~\citet{alsadik2025nbvreview}. We assume that all candidate images have already been captured and that reconstruction is itself a feed-forward operation. In this regime, selection methods that require iterative reconstruction can easily exceed the cost of the downstream task they are intended to accelerate. We therefore focus on a retrospective formulation: given a fixed pool of captured frames and a set of target viewpoints, select the subset that best supports reconstruction of those targets without rendering, fitting, or optimizing a scene during selection.~\citet{zhang2026peering} learn uncertainty-based active view selection without iterative reconstruction. Our selection-conditioned scorer is instead trained on jointly optimized subsets rather than per-view uncertainty estimates.

\textbf{Learning from combinatorial optimization.}
A related line of work learns to approximate the outputs of expensive combinatorial optimization procedures. Neural policies have been trained by imitation learning to approximate costly branch-and-bound decisions~\citep{gasse2019exact}, while ranking-based distillation transfers high-quality combinatorial policies into lightweight scoring models~\citep{woo2022efficient}. Our setting differs in that the optimizer produces high-quality reconstruction frame subsets, which we convert into ordered, selection-conditioned supervision for a sequential selector.


\vspace{-0.35em}

\section{Method}
\label{sec:method}
Our goal is to learn a scoring network $f_\theta$ that, given a pool of candidate frames and the poses of the views to be rendered, is applied greedily at inference time to select $K$ input frames without rendering, reconstruction, or per-scene optimization (\autoref{fig:pipeline}). Section~\ref{sec:setup} describes the problem and the three sets of views involved, \autoref{sec:features} the per-frame features, and \autoref{sec:selector} the scorer and the greedy selection loop that runs on them. The offline stage, run once on training scenes and never repeated for a new scene, comprises a genetic search that evaluates candidate subsets by reconstructing and rendering with them (\autoref{sec:ga}) and the behavior cloning that trains the scorer to reproduce its subsets one frame at a time (\autoref{sec:training}).

\subsection{Problem setup}
\label{sec:setup}
A capture sequence provides $N$ posed frames, which we partition into three disjoint sets: held-out target views $\gH$, optimization views $\gO$, and a candidate pool $\gP$. The target views $\gH$ are set aside first; each target pose is then matched to the remaining frame with the closest camera position and viewing direction (Euclidean distance between camera centers and angle between viewing directions, normalized per scene and weighted equally), and these matches form $\gO$. The remaining frames form the candidate pool $\gP$, from which a feed-forward reconstructor $R(\cdot)$ receives $K \ll |\gP|$ input views and renders $\gH$.
The optimal subset is:
\begin{equation}
\label{eq:optimal_subset}
S^\star = \argmax_{S \subseteq \gP,\; |S|=K} q(R(S),\gH),
\end{equation}
where $S$ denotes a subset of $K$ views selected from $\gP$, and $q$ measures reconstruction quality on the target views. Solving this optimization directly is impractical for two reasons. First, evaluating a subset requires running the reconstructor and rendering the target views, making the objective expensive to compute. Second, the contribution of a frame is highly context-dependent: the utility of a view depends on the other views selected alongside it, making $q$ non-additive over $S$.
During selection, the RGB images of the target views are unavailable, although their camera poses are known. Standard novel-view-synthesis splits give a selector no information about where rendering will happen; we assume instead that the target poses are known, as with a planned camera path, a robot trajectory, or a second capture pass. The optimization views are the mechanism that lets the selector use these poses without accessing the target images: they serve as observable proxies for the targets, and all target-related features (e.g., camera-center distance, viewing-direction angle, and appearance similarity) are computed between candidate frames and $\gO$. The target-view RGB images are never used during subset selection, training, or subset search; target poses are used solely to identify $\gO$.

We further test our method under a cross-capture protocol to show that it remains effective when the target views come from a separate capture and are spatially separated from the available frames. During training, a coverage level $p\in\{25,40,60,80,100\}$ controls the spatial spread: we randomly choose an anchor camera position, define the target region as the $p\%$ of capture frames closest to that anchor, and sample a coverage-dependent number of target views, so that $p=100$ means targets may be sampled anywhere in the capture. Labels are generated at each of the five coverage levels, and a single model is trained jointly across them.

\subsection{Features}
\label{sec:features}

Each candidate frame $i \in \gP$ is represented by a fixed-dimensional feature vector whose size is independent of both $|\gP|$ and $|\gO|$. 
We write $\vx_i(\gO,S) \in \R^{\ell}$ for the feature vector of candidate $i$ given the optimization views $\gO$ and the current selection $S$.
Features are divided into two relational groups and one intrinsic group:

\textbf{Target-relational features.} These features remain fixed throughout frame selection. They summarize the relationship between a candidate frame and the optimization views, characterizing its relevance for reconstructing the target region. They include statistics of camera-center distance, viewing-direction difference, appearance similarity computed as the cosine similarity between DINOv2~\citep{oquab2024dinov2} features, and the relative geometry of each candidate–optimization-view pair, including co-visibility.
\\
\textbf{Selection-conditioned features.} These features are recomputed after each selection step and capture the relationship between a candidate and the current subset. They characterize \emph{complementarity} with respect to the views already selected. They include the camera-center distance to the nearest selected frame, appearance similarity to the most similar selected frame, the minimum viewing-direction difference to any selected frame, and the angular novelty with respect to the nearest selected frame.
\\
\textbf{Intrinsic features.} We use image sharpness, measured as the variance of the Laplacian, to discourage the selection of blurry frames.

All features, except those constant across candidates, are standardized independently within each scene, reducing differences in scale and absolute magnitude. At the first selection step, all selection-conditioned features are set to zero; consequently, the initial choice depends only on target-relational features and image sharpness. The full feature specification is provided in Appendix~\ref{app:featconfig}.

\subsection{Scorer and inference}
\label{sec:selector}

The scorer $f_\theta$ is a multilayer perceptron with weights shared across all candidates.
Given the candidate pool $\gP$, it takes a $|\gP| \times \ell$ feature matrix as input and produces a $|\gP|$-dimensional score vector. Rather than modeling candidate interactions within the network, interactions are encoded explicitly through the selection-conditioned features. As a result, the same scorer can operate on candidate pools of arbitrary size while remaining permutation equivariant. Selection is greedy: starting from $S_0=\emptyset$, at each step $t=1,\dots,K$ the selector recomputes the features of every remaining candidate given the current selection, scores them, and appends the best one:

\begin{align}
s_i^{(t)} &= f_\theta\!\left(\vx_i(\gO,S_{t-1})\right),
  && i \in \gP\setminus S_{t-1}, \label{eq:score}\\
S_t &= S_{t-1}\cup\{i_t^\star\},
  && i_t^\star=\argmax_{i \in \gP\setminus S_{t-1}} s_i^{(t)}, \label{eq:inference}
\end{align}
and returns $S$ after $K$ steps.

In these equations, the feature vector $\vx_i(\gO,S_{t-1})$ is the only input to the scorer that depends on the current state of the selection $S_{t-1}$. The scoring function $f_\theta$ is applied to one candidate at a time and never sees the set.
Consequently, appending a frame to $S_{t-1}$ modifies only the selection-conditioned features, while all target-relational and intrinsic features remain unchanged. This design yields efficient inference: each of the $K$ selection steps performs a single forward pass through the scoring network, and no rendering, reconstruction, or per-scene optimization is required.

\subsection{Supervision by subset search}
\label{sec:ga}

The objective in \autoref{eq:optimal_subset} is both combinatorial and expensive to evaluate. Exhaustive search requires assessing $\binom{|\gP|}{K}$ candidate subsets, and each evaluation entails reconstruction and rendering.
We therefore approximate high-quality solutions offline using a genetic algorithm. Each individual represents a $K$-view subset of the candidate pool. To encourage diversity, the initial population, consisting of $C$ individuals (we use $C=40$), combines farthest-view sampling from multiple anchors, uniformly spaced sampling at different phase offsets, and random subsets. This initialization provides a broad set of geometrically plausible solutions rather than relying on a single heuristic. Since the target images are unavailable, each individual is evaluated by reconstructing from its views and rendering the optimization views: the search optimizes $q(R(S), \gO)$ as a proxy for~\autoref{eq:optimal_subset}. Fitness is defined as the average of PSNR, SSIM~\citep{wang2004ssim}, and negated LPIPS~\citep{zhang2018lpips}, with each metric min-max normalized across the evaluated individuals for a scene to prevent any single metric from dominating the objective.

The population is evolved for $G=15$ generations. At each generation, parents are chosen by tournament selection: a small group of individuals is sampled, and the fittest is retained. Crossover forms a new subset by drawing $K$ distinct views from the union of two parents, while mutation replaces a few views with randomly sampled candidates to maintain exploration; both operators preserve the cardinality $K$.  
The fittest individuals are copied unchanged to the next generation. Since fitness is measured through reconstruction and rendering, the resulting subsets provide supervision directly aligned with the reconstruction objective. Its computational cost, however, makes it unsuitable for deployment: hundreds of candidate subsets may need to be reconstructed and rendered for a single scene, whereas the downstream reconstructor operates in a single forward pass. Rather than retaining only the highest-scoring subset, we keep the top 4 solutions for each scene and use them as supervision during training, as described in the next section. This exposes the selector to multiple high-quality view configurations, reflecting the inherent non-uniqueness of the subset-selection problem, and consistently improves generalization in our experiments. Implementation details, hyperparameters and training data are given in Appendix~\ref{app:ga_details} and Appendix~\ref{app:training_details}.

\subsection{Training}
\label{sec:training}

The search of \autoref{sec:ga} returns sets of frames; the scorer must learn from them one decision at a time.
We also experimented with a single-pass formulation,  in which the scorer jointly scores all frames, and the $K$ frames with the highest scores are selected. This led to generalization problems when considering different values of $K$ at inference time.
%
We instead predict scores iteratively: given the frames already selected (called the prefix), the scorer is trained so that the \emph{next} frame of the subset receives the highest score, with the selection-conditioned features of \autoref{sec:features} carrying the dependence on the current prefix. This replaces one global, scene-specific decision by $K$ local ones. This lets the selector generalize across scenes, budget $K$, and pool size.
This is behavior cloning of an expensive expert that makes discrete decisions, e.g.,~\citep{gasse2019exact}. In our case, the expert is the combinatorial optimization together with an ordering rule. The selector learns to reproduce its decisions one step at a time. It takes two steps: first, order each retained set into a sequence of per-step labels; second, build the per-step target probabilities from these sequences and train the scorer to match them.

\textbf{Step 1: label ordering.}
Any ordering of a retained subset $S$ yields the same final selection, but the ordering determines what each per-step label teaches~\citep{vinyals2015order}. We therefore order $S$ into a list $\Pi_{K}=[\pi_1,\dots,\pi_{K}]$ so that the induced trajectory is consistent with the incremental structure the selector must learn: the first frame is useful in isolation, and later ones' value lies in what they add to the prefix. Starting from the empty list, we repeatedly append the remaining frame of $S$ that most reduces the residual distance from the optimization views to their nearest selected frame, 
\begin{equation}
   \pi_t = \argmin_{i \,\in\, S \setminus \Pi_{t-1}}
       \sum_{o \in \gO} \min\!\left(c_{t-1}(o),\, d(i,o)\right),
   \label{eq:order}
\end{equation}
where $t=1,\dots,K$ indexes the position, $d(i,o)$ is the scene-normalized distance between the camera centers of candidate frame $i$ and optimization view $o$, $\Pi_{t-1}=[\pi_1,\dots,\pi_{t-1}]$ is the prefix of the ordering with $\Pi_0=\emptyset$, and $c_{t-1}(o)=\min_{i \in \Pi_{t-1}} d(i,o)$ is the distance from $o$ to its nearest frame in the prefix, with $c_0(o)=\infty$. This construction makes the marginal gain of each label explicit at the step where it is supervised, matching what the selection-conditioned features can express. It is internal to training: it converts a set label into $K$ per-step labels and leaves the subset unchanged.

\textbf{Step 2: per-step targets and objective.}
Training mirrors inference. Each retained subset $S$ defines one trajectory $[\pi_1,\dots,\pi_K]$ through \autoref{eq:order}, and we write $\Ls(S)$ for its loss: starting from an empty selection, the model scores all remaining candidates at each position and is trained to predict the next frame of the trajectory, with previously selected frames masked out. The scores of the remaining candidates are normalized into probabilities with a softmax, the target assigns probability 1 to the next frame of the trajectory (and 0 to all others), and we use cross-entropy:

\begin{equation}
\label{eq:loss}
\Ls(S) = -\frac{1}{K}\sum_{t=1}^{K} \log p^{(t)}_{\pi_t},
\qquad
p^{(t)}_i = \frac{\exp\!\big(s^{(t)}_i\big)}
                 {\sum_{i' \in \gP \setminus \Pi_{t-1}} \exp\!\big(s^{(t)}_{i'}\big)},
\end{equation}
where $s^{(t)}_i = f_\theta\!\left(\vx_i(\gO,\Pi_{t-1})\right)$ is the score of candidate $i$ at position $t$, computed as in \autoref{eq:score} with the search prefix $\Pi_{t-1}$ in place of the model's own selection $S_{t-1}$. 
In \autoref{eq:loss}, $p_{\pi_t}^{(t)}$ is the probability assigned by the selector to the frame $\pi_t$ prescribed by the ordering criterion; minimizing the loss encourages this frame to receive the highest probability among the remaining candidates.
Before each position, the selection-conditioned features are recomputed against $\Pi_{t-1}$, while the target-relational and intrinsic features are computed once per scene. The per-step losses are averaged over the $K$ positions and applied as a single update, so one training example is one complete trajectory for a (scene, subset) pair. 

The search does not return a unique optimum: distinct subsets can cover the target views in geometrically different but equally effective ways, and supervising on a single maximum score would ask the selector to discriminate between solutions the objective itself does not distinguish. We therefore retain the top $M=4$ subsets $S^{(1)},\dots,S^{(M)}$ per scene and weight their trajectories by fitness:
\begin{equation}
\label{eq:weight}
\tilde w_m = \exp\!\left(
\frac{F_m-\max_{m'}F_{m'}}
{\sigma_F+\eps}
\right),
\qquad
w_m =
\frac{\tilde w_m}
{\tfrac{1}{M}\sum_{m'}\tilde w_{m'}},
\qquad
\Ls_m = w_m\,\Ls\big(S^{(m)}\big)
\end{equation}
where $F_m$ is the fitness of $S^{(m)}$ as defined in \autoref{sec:ga}, $\sigma_F$ is the standard deviation of the $M$ retained fitnesses, and $\eps$ is a small constant for numerical stability. Setting the temperature to $\sigma_F$ normalizes the scale of fitness differences. Normalizing the weights to mean one keeps the loss scale independent of $M$. The selector is thus fitted to the local distribution of high-quality solutions rather than to one search outcome, with lower-ranked subsets acting as regularization. The total loss is the average of $\Ls_m$ over the retained subsets and training scenes.



\vspace{-0.35em}

\section{Results}
\label{sec:exp}


Every capture is split into a candidate pool $\gP$, optimization views $\gO$ and held-out target views $\gH$ as described in Section~\ref{sec:setup}; $\gH$ is used only for evaluation. The split is deterministic and shared by every method, so all selections are scored on identical held-out views of identical scenes by the same reconstructor, which, unless stated otherwise, is LongLRM~\citep{ziwen2025long} (Appendix~\ref{app:reconstructor}). For transfer experiments, we additionally
evaluate the novel view synthesis using Nerfstudio~\citep{tancik2023nerfstudio} based on NeRF~\citep{mildenhall2020nerf}, and 3D Gaussian Splatting (3DGS)~\citep{kerbl20233dgs}. We report
PSNR\,/\,SSIM~\citep{wang2004ssim}\,/\,LPIPS~\citep{zhang2018lpips} averaged over scenes.
\\
\textbf{Datasets.} Our comparison covers six corpora spanning outdoor captures, unbounded scenes and handheld
indoor video: DL3DV~\citep{ling2024dl3dv}, Tanks and Temples~\citep{knapitsch2017tanks},
Mip-NeRF~360~\citep{barron2022mipnerf360}, ScanNet-iPhone and
ScanNet++~\citep{yeshwanth2023scannetpp}, and 7-Scenes~\citep{shotton2013scene}. The selector is trained on DL3DV labels only, with feature selection performed on a disjoint ScanNet++ validation set; every other dataset is
therefore a zero-shot transfer, and DL3DV evaluation scenes are held out from training-label generation and training.
\\
\textbf{Baselines.} We compare against spread-based sampling (random, uniform, farthest-view sampling),
photogrammetric selection (NeRF-Director~\citep{xiao2024nerfdirector}, using co-visibility over shared sparse
points), the MVSNet view-selection score (MVSNet~\citep{yao2018mvsnet}), set-aware non-learned criteria (DPP, a greedy
determinantal-point-process log-determinant over appearance similarity~\citep{kulesza2012dpp}, and $k$-medoids~\citep{kaufman2009finding}
on joint pose and appearance),
coverage-based view selection (COVER~\citep{chen2026cover}), information-gain selection (FisherRF~\citep{jiang2024fisherrf}), which is far more expensive and is therefore evaluated only in the six-dataset comparison. 
Baselines may select from both the candidate pool $\gP$ and the optimization views $\gO$, whereas our method selects only from $\gP$.

\subsection{Selection quality and computational efficiency}
\label{sec:exp_lessismore}

\begin{figure*}[t]
    \centering
    \includegraphics[width=0.88\textwidth]{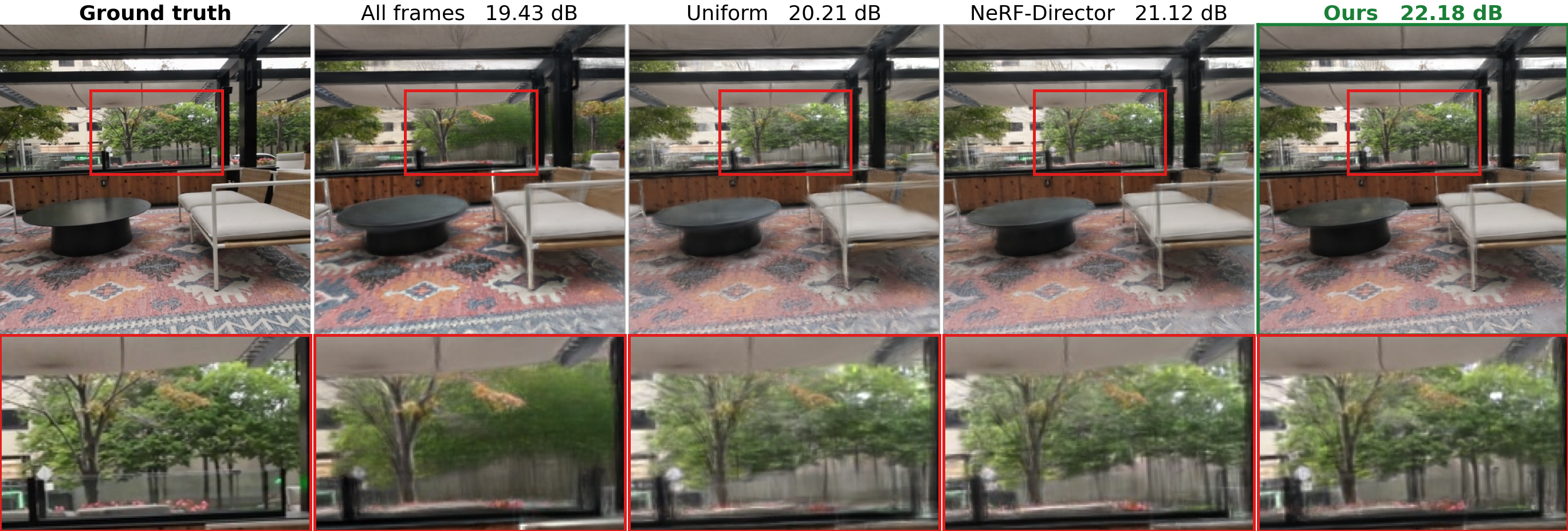}
    \caption{\textbf{Novel-view reconstructions on DL3DV}, $K{=}40$; further datasets in Appendix~\ref{app:qualitative}.}
    \label{fig:qual_dl3dv}
\end{figure*}

Our learned selector outperforms in PSNR the strongest baseline across all six datasets and every evaluated frame budget (Figures~\ref{fig:heatmap} and~\ref{fig:qual_dl3dv}).
On DL3DV, using $30$ to $50$ selected frames improves both reconstruction quality and computational efficiency relative to using the full sequence 
(Table~\ref{tab:time_quality}). At $K{=}30$ and $K{=}40$, our selected subsets improve PSNR by $1.33$ and $1.68$\,dB, respectively, while reducing reconstruction time by factors of $7.3$ and $5.1$. Peak reconstruction memory also drops from $17.8$\,GB for the full sequence to $4.6$\,GB at $K{=}30$ and $5.9$\,GB at $K{=}40$.
Feature ablations further reduce selection time with a small decrease in reconstruction quality. At $K{=}30$, removing appearance features reduces selection time from $0.98$ to $0.10$\,s, while PSNR decreases by only $0.03$\,dB, from $22.33$ to $22.30$\,dB. Additionally omitting sharpness reduces selection time to $0.04$\,s, with PSNR remaining at $22.02$\,dB.
Complete results for all datasets and baselines are provided in Appendix~\ref{app:budget_sweep}.

\begin{table*}[t]
\centering
\scriptsize
\setlength{\tabcolsep}{2.5pt}
\renewcommand{\arraystretch}{0.63}
\begin{minipage}[c]{0.47\linewidth}
\centering

\captionof{table}{\textbf{Cross-capture novel view synthesis} on 7-Scenes.}
\label{tab:crossvideowopt}

\vspace{0.3em}

\resizebox{\linewidth}{!}{%
\begin{tabular}{lccc@{\hskip 1.2em}ccc}
\toprule
& \multicolumn{3}{c}{$K{=}30$} & \multicolumn{3}{c}{$K{=}40$} \\
\cmidrule(lr){2-4}\cmidrule(lr){5-7}
Method & PSNR$\uparrow$ & SSIM$\uparrow$ & LPIPS$\downarrow$
       & PSNR$\uparrow$ & SSIM$\uparrow$ & LPIPS$\downarrow$ \\

\midrule
Random                 & $18.57$ & $0.654$ & $0.342$ & $18.92$ & $0.662$ & $0.317$ \\
Uniform                & $19.12$ & $0.668$ & $0.310$ & $19.24$ & $0.670$ & $0.299$ \\
FVS          & $18.81$ & $0.659$ & $0.326$ & $19.19$ & $0.666$ & $0.306$ \\
NeRF-Director          & $18.82$ & $0.659$ & $0.324$ & $19.20$ & $0.667$ & $0.306$ \\
DPP                    & $18.75$ & $0.656$ & $0.326$ & $19.12$ & $0.662$ & $0.308$ \\
$k$-medoids            & $19.19$ & $0.666$ & $0.306$ & $19.40$ & $0.670$ & $0.294$ \\
COVER                  & $17.77$ & $0.636$ & $0.361$ & $18.58$ & $0.653$ & $0.328$ \\
\midrule
\textbf{Ours}                  & $\mathbf{19.28}$ & $\mathbf{0.677}$ & $\mathbf{0.296}$ & $\mathbf{19.49}$ & $\mathbf{0.680}$ & $\mathbf{0.290}$ \\

\bottomrule
\end{tabular}%
}

\end{minipage}
\hfill
\begin{minipage}[c]{0.47\linewidth}

\centering
\small
\caption{\textbf{Object-centric novel view synthesis} on ScanNet~v2 and CO3D, $K{=}30$.}
\label{tab:objectcentric}
\vspace{0.3em}
\resizebox{\linewidth}{!}{%
\begin{tabular}{l ccc ccc}
\toprule
& \multicolumn{3}{c}{\textbf{ScanNet~v2} ($n{=}10$)} & \multicolumn{3}{c}{\textbf{CO3D} ($n{=}10$)} \\
\cmidrule(lr){2-4}\cmidrule(lr){5-7}
Selection & PSNR$\uparrow$ & SSIM$\uparrow$ & LPIPS$\downarrow$ & PSNR$\uparrow$ & SSIM$\uparrow$ & LPIPS$\downarrow$ \\
\midrule
Random & $20.69$ & $0.684$ & $0.316$ & $23.56$ & $0.816$ & $0.209$ \\
Uniform & $21.12$ & $0.701$ & $0.281$ & $23.93$ & $0.826$ & $0.200$ \\
FVS & $21.13$ & $0.694$ & $0.297$ & $23.72$ & $0.820$ & $0.207$ \\
NeRF-Director & $21.54$ & $0.703$ & $0.281$ & $23.71$ & $0.821$ & $0.207$ \\
DPP & $20.93$ & $0.692$ & $0.303$ & $23.78$ & $0.822$ & $0.209$ \\
$k$-medoids & $20.80$ & $0.694$ & $0.289$ & $23.90$ & $0.818$ & $0.210$ \\
COVER & $20.58$ & $0.677$ & $0.313$ & $23.41$ & $0.814$ & $0.218$ \\
\midrule
\textbf{Ours} & $\mathbf{22.04}$ & $\mathbf{0.714}$ & $\mathbf{0.265}$ & $\mathbf{24.75}$ & $\mathbf{0.847}$ & $\mathbf{0.175}$ \\
\bottomrule
\end{tabular}}
\end{minipage}

\end{table*}

\subsection{Generalization across reconstructors and paradigms}
\label{sec:exp_transfer}

\textbf{Reconstructor-agnostic transfer.} The selector is supervised with labels from one reconstructor; if what it learns is a property of good input sets rather than of that model, the same frozen weights should help elsewhere.
Table~\ref{tab:transferwopt} applies the same frozen weights to Nerfstudio and 3DGS under an
identical protocol with same scenes, candidate pools, and held-out targets, only the reconstructor changes. These results support generalization across the different reconstruction paradigms.

\begin{table}[t]
\centering
\small
\caption{\textbf{Cross-reconstructor transfer of selected views.} Results of novel view synthesis on DL3DV.}  
\label{tab:transferwopt}
\vspace{0.2em}
\setlength{\tabcolsep}{2.5pt}
\renewcommand{\arraystretch}{0.63}
\resizebox{\linewidth}{!}{%
\begin{tabular}{l ccc ccc}
\toprule
& \multicolumn{3}{c}{$K=30$} & \multicolumn{3}{c}{$K=40$} \\
\cmidrule(lr){2-4}\cmidrule(lr){5-7}
Selector
& LongLRM & Nerfstudio & 3DGS
& LongLRM & Nerfstudio & 3DGS \\
& PSNR$\uparrow$ / SSIM$\uparrow$ / LPIPS$\downarrow$ & PSNR$\uparrow$ / SSIM$\uparrow$/ LPIPS$\downarrow$ & PSNR$\uparrow$ / SSIM$\uparrow$ / LPIPS$\downarrow$
& PSNR$\uparrow$ / SSIM$\uparrow$ / LPIPS$\downarrow$ & PSNR$\uparrow$ / SSIM$\uparrow$ / LPIPS$\downarrow$ & PSNR$\uparrow$ / SSIM$\uparrow$ / LPIPS$\downarrow$ \\
\midrule
Random
& $21.14 / 0.686 / 0.213$
& $18.96 / 0.571 / 0.427$
& $18.57 / 0.635 / 0.263$
& $21.72 / 0.701 / 0.197$
& $19.65 / 0.602 / 0.393$
& $19.80 / 0.676 / 0.237$ \\

Uniform
& $21.60 / 0.696 / 0.201$
& $19.23 / 0.583 / 0.418$
& $19.44 / 0.667 / 0.235$
& $22.32 / 0.719 / 0.181$
& $20.16 / 0.625 / 0.372$
& $20.61 / 0.708 / 0.213$ \\

FVS
& $21.99 / 0.711 / 0.187$
& $19.03 / 0.569 / 0.420$
& $19.65 / 0.675 / 0.231$
& $22.51 / 0.727 / 0.174$
& $20.09 / 0.618 / 0.361$
& $20.93 / 0.717 / 0.208$ \\

NeRF-Director
& $21.75 / 0.701 / 0.191$
& $18.97 / 0.567 / 0.427$
& $19.42 / 0.664 / 0.238$
& $22.25 / 0.716 / 0.177$
& $19.58 / 0.594 / 0.396$
& $20.37 / 0.698 / 0.225$ \\

DPP
& $21.59 / 0.697 / 0.197$
& $18.98 / 0.569 / 0.426$
& $19.14 / 0.652 / 0.250$
& $22.19 / 0.716 / 0.180$
& $19.73 / 0.600 / 0.397$
& $20.29 / 0.696 / 0.224$ \\

$k$-medoids
& $21.68 / 0.699 / 0.194$
& $19.10 / 0.576 / 0.416$
& $19.28 / 0.659 / 0.247$
& $22.26 / 0.718 / 0.179$
& $20.06 / 0.620 / 0.374$
& $20.35 / 0.701 / 0.222$ \\

COVER
& $21.59 / 0.706 / 0.204$
& $19.38 / 0.592 / 0.395$
& $20.21 / 0.690 / 0.216$
& $22.34 / 0.724 / 0.188$
& $20.34 / \mathbf{0.638} / \mathbf{0.331}$
& $20.96 / 0.718 / 0.203$ \\
\midrule
\textbf{Ours}
& $\mathbf{22.33} / \mathbf{0.724} / \mathbf{0.177}$
& $\mathbf{19.92} / \mathbf{0.598} / \mathbf{0.372}$
& $\mathbf{21.44} / \mathbf{0.732} / \mathbf{0.178}$
& $\mathbf{22.68} / \mathbf{0.734} / \mathbf{0.171}$
& $\mathbf{20.52} / 0.634 / 0.337$
& $\mathbf{22.22} / \mathbf{0.756} / \mathbf{0.173}$ \\

\bottomrule
\end{tabular}}
\end{table}

\textbf{Object centric novel view synthesis.} We evaluate the same frozen selector on the reconstruction of a designated object evaluating on 10 scenes each from ScanNet~v2~\citep{dairichly}  and CO3D~\citep{reizenstein21co3d} (\autoref{tab:objectcentric}). ScanNet~v2 contains objects in cluttered rooms; CO3D depicts individual objects. Both provide ground-truth masks for the objects.  Compared with the strongest baseline, our method improves PSNR by $0.50$\,dB on ScanNet~v2 and $0.82$\,dB on CO3D.

\begin{figure*}[t]
    \centering
    \includegraphics[width=0.88\textwidth]{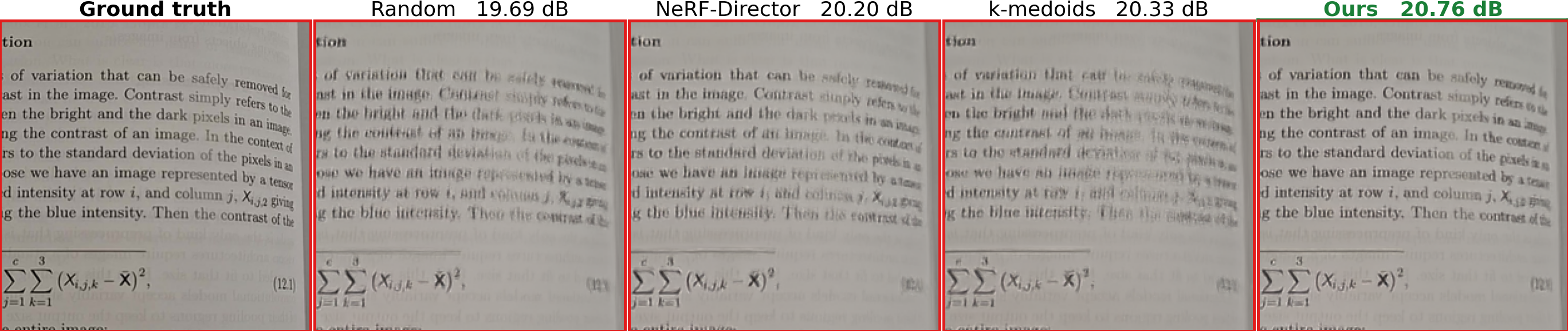}
    \caption{\textbf{Real-world cross-capture novel view synthesis}, $K{=}30$.}
    \label{fig:real_world}
\end{figure*}

\textbf{Cross-capture novel view synthesis.}
Every protocol so far holds out targets from \emph{inside} the candidate pool's own trajectory, so each target has temporal neighbors among the selectable frames. Deployment, however, may require synthesizing views at unobserved poses from frames of a separate capture. We therefore evaluate a second protocol in which the target views come from one held-out capture of a scene and every candidate and optimization view comes from the \emph{other} captures: no frame of the target capture is selectable, so the reconstruction is assembled from separate passes and scored on a pass it has never seen. 7-Scenes supports this directly, since each room is recorded as several independent videos with a common pose frame; we follow the dataset's sequence split, report one measurement per (room, held-out video) pair, and retain every fifth frame of each video so that even the most expensive baselines run in reasonable time. Targets are drawn from the half of the held-out video closest to the remaining captures.
Under this cross-capture protocol, our method is ahead of every baseline at both frame budgets (Table~\ref{tab:crossvideowopt}). For the six rooms involved, reconstructor and selector weights remain unchanged; only the division into target and selectable captures differs. Figure~\ref{fig:real_world} shows the same setting on a real-world capture, where the selected subset yields a more faithful reconstruction from observations acquired along a separate trajectory.


\subsection{Ablations}
\label{sec:exp_ablations}

\textbf{Pose estimation.}
Our selector does not require ground-truth camera poses. Table~\ref{tab:vggtposes} replaces the COLMAP~\cite{schoenberger2016sfm} poses used throughout the main experiments with poses estimated by VGGT-$\Omega$~\cite{wang2026vggtomega} on DL3DV, while leaving the selector and reconstruction pipeline unchanged. Performance decreases under the estimated poses, with the gap becoming more visible at larger input budgets, but the selector remains effective across the full budget range. This indicates that accurate geometry benefits selection, while the method does not depend on a particular pose-estimation pipeline and remains usable when poses must be inferred directly from the input sequence.

\textbf{Feature representation.}
Table~\ref{tab:time_quality} ablates the appearance and sharpness components of the deployed selector, while keeping its architecture and training procedure fixed. Removing appearance features produces only a small reduction in reconstruction quality across budgets, while substantially reducing selection time. Removing sharpness in addition to appearance leads to a clearer and consistently larger degradation, despite making inference cheaper. Most of the selector's performance thus comes from the inexpensive geometric and selection-context features; appearance adds a smaller complementary gain, and sharpness, although extremely cheap to compute, carries information not captured by geometry alone. Further details on feature representations and combinations are provided in Appendix~\ref{app:featconfig} and Appendix~\ref{app:ablations}.

\begin{table*}[t]
\setlength{\abovecaptionskip}{0pt}
\small

\noindent\makebox[\textwidth][s]{
\setlength{\tabcolsep}{2.5pt}
\renewcommand{\arraystretch}{0.93}
\begin{minipage}[t]{0.31\textwidth}
\vspace{0pt}
\centering
\captionof{table}{\textbf{Robustness to estimated poses.} Reconstruction quality on DL3DV using COLMAP and VGGT-$\Omega$ camera poses across input budgets.}
\label{tab:vggtposes}
\resizebox{0.65\linewidth}{!}{%
\begin{tabular}{lcc}
\toprule
& \multicolumn{2}{c}{\textbf{Ours}} \\
& \multicolumn{2}{c}{\scriptsize PSNR$\uparrow$/SSIM$\uparrow$/LPIPS$\downarrow$} \\
\cmidrule(lr){2-3}
$K$ & COLMAP & VGGT-$\Omega$ \\
\midrule
10 & 15.69/.571/.349 & 15.48/.536/.366 \\
20 & 20.97/.696/.204 & 19.95/.629/.231 \\
30 & 22.33/.724/.177 & 20.89/.648/.209 \\
40 & 22.68/.734/.171 & 21.11/.653/.202 \\
50 & 22.76/.738/.168 & 21.07/.654/.201 \\
\bottomrule

\end{tabular}%
}

\end{minipage}
\hfill
\begin{minipage}[t]{0.646\textwidth}
\vspace{0pt}
\centering
\setlength{\tabcolsep}{2.5pt}
\renewcommand{\arraystretch}{0.38}
\captionof{table}{\textbf{Quality, runtime and memory per scene on DL3DV.} Reconstruction time and peak
memory are shared by all variants, since each feeds $K$ views to the same reconstructor. The last row feeds the whole candidate pool, $K = |\gP|$.}
\label{tab:time_quality}
\resizebox{\linewidth}{!}{%
\begin{tabular}{l ccc ccc c ccc c}
\toprule
& \multicolumn{3}{c}{\textit{PSNR}~$\uparrow$}
& \multicolumn{3}{c}{\textit{Selection time} (s)~$\downarrow$}
& \multirow{2}{*}{\shortstack{\textit{Recon.}\\(s)~$\downarrow$}}
& \multicolumn{3}{c}{\textit{Selec. + Recon. time} (s)~$\downarrow$}
& \multirow{2}{*}{\shortstack{\textit{Peak mem.}\\(GB)~$\downarrow$}} \\
\cmidrule(lr){2-4}
\cmidrule(lr){5-7}
\cmidrule(lr){9-11}

$K$
& \textbf{Ours} & \shortstack{w/o\\appear.} & \shortstack{w/o appear.\\\& sharp.}
& \textbf{Ours} & \shortstack{w/o\\appear.} & \shortstack{w/o appear.\\\& sharp.}
&
& \textbf{Ours} & \shortstack{w/o\\appear.} & \shortstack{w/o appear.\\\& sharp.}
& \\
\midrule

10 & \textbf{15.69} & 15.69 & 15.47
   & 0.95 & 0.07 & \textbf{0.02}
   & 0.19
   & 1.14 & 0.26 & \textbf{0.21} & 2.0 \\

20 & \textbf{20.97} & 20.85 & 20.63
   & 0.98 & 0.09 & \textbf{0.03}
   & 0.26
   & 1.24 & 0.35 & \textbf{0.29} & 3.3 \\

30 & \textbf{22.33} & 22.30 & 22.02
   & 0.98 & 0.10 & \textbf{0.04}
   & 0.39
   & 1.37 & 0.49 & \textbf{0.43} & 4.6 \\

40 & \textbf{22.68} & 22.66 & 22.44
   & 0.99 & 0.11 & \textbf{0.06}
   & 0.56
   & 1.55 & 0.67 & \textbf{0.62} & 5.9 \\

50 & \textbf{22.76} & 22.72 & 22.49
   & 1.00 & 0.13 & \textbf{0.07}
   & 0.75
   & 1.75 & 0.88 & \textbf{0.82} & 7.2 \\

\midrule
All & 21.00 & -- & --
    & --  & -- & --
    & 2.85
    & \multicolumn{3}{c}{2.85} & 17.8 \\
\bottomrule
\end{tabular}%
}

\end{minipage}%
}

\end{table*}

\begin{figure*}[t]
    \centering
    \includegraphics[width=0.94\textwidth,
        height=3.9cm]{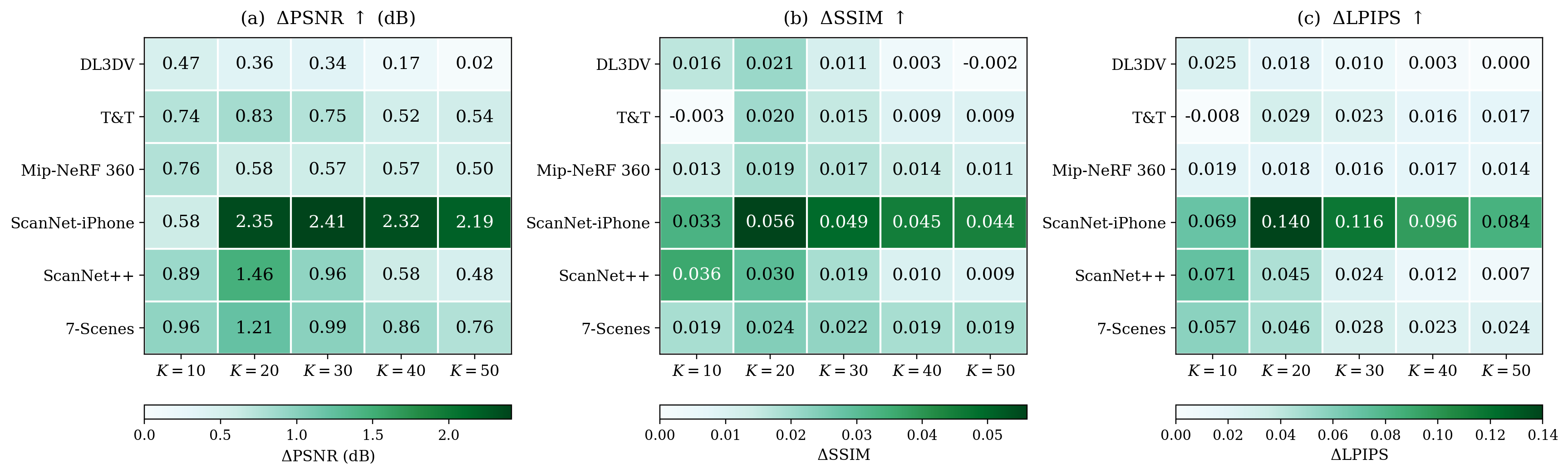}
    \caption{\textbf{Performance gain over the strongest baseline.} Improvement of our selector against the best baseline across datasets and number of frames.}
    \label{fig:heatmap}
\end{figure*}

\vspace{-0.35em}
\section{Conclusion}
\label{sec:conclusion}

Frame selection directly determines reconstruction quality in feed-forward novel view synthesis. Our approach distills an expensive, reconstruction-aware genetic search into a lightweight selector requiring no rendering, reconstruction, or per-scene optimization at inference time. Across diverse datasets and input budgets, the selector outperforms geometric, retrieval-based, and set-aware baselines, transfers without retraining to Nerfstudio and 3DGS, and extends to object-targeted selection. Ablations identify geometry and selection context as the main signals, while appearance offers dataset-dependent gains at additional computational cost.
Our approach involves two practical considerations. Optimization views provide a dataset- and geometry-dependent approximation to the evaluation objective. Additionally, both the selector and reconstructor require camera poses, which must be estimated beforehand if unavailable. These findings establish frame selection as an integral part of reconstruction, with target viewpoints guiding the choice of inputs. Future work could compute the
geometric features directly against the requested target poses, keeping the proxy views only for
appearance. Further directions include dynamic scenes, learned candidate features, and reinforcement-learning policies that optimize subsets directly.

\vspace{-0.55em}

\section*{AI use statement}

We used a generative AI coding assistant extensively throughout this project, and disclose its use across
the following categories.

\textbf{Tasks with required disclosure.} We used generative AI tools to \emph{implement} the following methods: the
selector, its feature extraction, the genetic-algorithm label generation, the baseline implementations, and
the evaluation harnesses. We used them to \emph{provide
feedback on research methodology and experiments}, including the target-coverage protocol and the object-targeted evaluation protocol. We used them to \emph{clean and reformat data}, and convert captures between dataset layouts. We used them to
\emph{diagnose} experimental anomalies. We used them to \emph{propose and refine hypotheses} concerning the source of the
observed gains, several of which we subsequently tested and rejected.

We did \textbf{not} use generative AI tools to generate synthetic datasets, to develop theoretical models or
conceptual frameworks, to formulate mathematical claims, or to translate text. Support for qualitative and thematic data analysis is not applicable to this work.

\textbf{Tasks with recommended disclosure.} We additionally used generative AI tools to create and edit
software code, to draft some parts of this paper, to identify, summarize, and analyze related literature and to
format references, to suggest the structure of the paper and propose its title, to suggest experimental
parameters, and for brainstorming and information search.

\textbf{Verification.} All experimental code was executed and its outputs inspected by the authors. Reported
numbers were regenerated from stored per-scene results, and comparisons between methods were recomputed on
paired scene sets after several unpaired aggregations were identified and corrected during the project.
Baseline implementations were checked against their original sources. All claims, citations and text in this
paper were reviewed by the authors, and we take responsibility for the final content of this work, including
text, claims and artifacts produced with the aid of generative AI.

\section*{Reproducibility statement}
The feature representation is specified in full in \autoref{app:featconfig}, the genetic search that
produces the training labels in \autoref{app:ga_details}, and the training data, splits and
hyperparameters in \autoref{app:training_details}. \autoref{app:reconstructor} reports the choice of
reconstructor, and \autoref{app:budget_sweep} the complete per-dataset, per-budget results behind the
summary figures. All benchmark experiments use public datasets with the splits described in \autoref{sec:exp}. We train our selector using three random seeds
and report its mean performance across these runs.

\bibliography{iclr2027_conference}
\bibliographystyle{iclr2027_conference}

\clearpage
\appendix
\begin{center}
{\Large\bfseries Appendix: Supplementary Material}\\[0.4em]
\end{center}
\vspace{0.5em}

\section{Additional Method Details}
\label{app:method_details}

\subsection{Feature representation}
\label{app:featconfig}

Each candidate frame is represented by 18 scalar features: 13 describe its relation to the proxy views, 4 describe its contribution relative to the currently selected frames, and one measures image sharpness. The four features involving the selected set are recomputed after each selection step. Candidate scores therefore depend on the frames already selected: a frame that initially provides useful coverage may become redundant once another frame with similar coverage has been selected.

\textbf{Relation to the optimization views.}
The optimization views are observed frames used to represent the target region. A camera position is its location in 3D space, and its viewing direction is the direction of its optical axis. The superscripts $\gO$ and $S$ distinguish comparisons with optimization views and already selected views, respectively.

For each candidate frame, we first compare its camera position, viewing direction and image appearance with those of every optimization view. The following eight features summarize these comparisons:
\begin{itemize}\setlength{\itemsep}{1pt}
  \item $d_{\min}^{\gO}$: Distance from the candidate camera position to the nearest optimization camera position. A small value indicates that the candidate was captured near at least one optimization view.

  \item $d_{\mathrm{mean}}^{\gO}$: Average distance from the candidate camera position to all optimization camera positions. This describes how close the candidate is to the optimization views as a group.

  \item $d_{25}^{\gO}$: The 25th percentile of the distances from the candidate camera position to the optimization camera positions. Approximately one quarter of the optimization cameras lie within this distance, so the feature describes proximity beyond the single nearest camera.

  \item $\gamma_{\min}^{\gO}$: Smallest angle between the candidate camera's viewing direction and the viewing direction of any optimization camera. A value near zero means that the candidate camera points in approximately the same direction as at least one optimization camera.

  \item $\gamma_{\mathrm{mean}}^{\gO}$: Average angle between the candidate camera's viewing direction and the viewing directions of all optimization cameras. This summarizes its directional alignment with the optimization views as a group.

  \item $\rho_{\max}^{\gO}$: Highest DINO~\cite{oquab2024dinov2} feature similarity between the candidate image and any optimization image. A high value indicates that the candidate resembles at least one optimization image in DINO feature space.

  \item $\rho_{\mathrm{mean}}^{\gO}$: Average DINO feature similarity between the candidate image and all optimization images. This describes its overall appearance similarity to the optimization views.

  \item $\sigma_{\rho}$: Standard deviation of the DINO similarity scores between the candidate image and the optimization images. A high value indicates that the candidate resembles some optimization images much more closely than others; a low value indicates that these similarity scores vary little.
\end{itemize}

The following features describe the relative geometry between a candidate and the optimization views. We assign one reference point to each optimization view. Starting from the optimization camera center, we follow its viewing direction and choose the point along this ray that passes closest to the centroid of all candidate-camera positions. These reference points are computed once and remain fixed throughout selection. To compare a candidate with an optimization view, we use the reference point assigned to that optimization view. The triangulation angle is the angle between the direction from this point to the candidate camera and the direction from this point to the optimization camera. We compute one such angle for every candidate-optimization pair. Unlike $\gamma^{\gO}$, this angle compares the camera positions relative to the reference point rather than the directions in which the cameras point.

\begin{itemize}\setlength{\itemsep}{1pt}

    \item $\tau_{\mathrm{tri}}$: For every candidate-optimization pair, we evaluate its triangulation angle using the view-selection score adopted by MVSNet~\cite{yao2018mvsnet}. The score is low for very small or very large angles and highest for an intermediate angle. We define $\tau_{\mathrm{tri}}$ as the highest score obtained across all optimization views.

  \item $\alpha_{\mathrm{mean}}$: Average triangulation angle over all candidate-optimization pairs. It summarizes how differently the candidate and optimization cameras are positioned around their reference locations.

\item $b_{\perp}$: For each optimization view, we measure the shortest distance from the candidate camera center to the line along which the optimization camera points. The feature is the average of these distances over all optimization views.

  \item $b_{\parallel}$: Average forward or backward displacement of the candidate camera along the viewing directions of the optimization cameras.

  \item $c_{\mathrm{vis}}$: Fraction of optimization-view reference point locations lying within the candidate camera's viewing cone. A high value means that the candidate points toward the locations represented by many optimization views. This is a pose-based visibility indicator and does not account for occlusion.
\end{itemize}

\textbf{Relation to the selected frames.}
These features compare a candidate with the frames already selected. They are recomputed after each addition because a candidate's contribution depends on the current selection:
\begin{itemize}\setlength{\itemsep}{1pt}

  \item $d_{\min}^{S}$: Distance from the candidate camera position to the nearest selected camera position. A small value indicates that the current selection already contains a camera close to the candidate.

  \item $\gamma_{\min}^{S}$: Smallest angle between the candidate camera's viewing direction and that of any selected camera. A value near zero indicates that at least one selected camera already points in approximately the same direction.

  \item $\rho_{\max}^{S}$: Highest DINO feature similarity between the candidate image and any selected image. A high value indicates that the current selection already contains an image with similar appearance in DINO feature space.

  \item $\phi_{\mathrm{nov}}$: Angular separation between the directions from a common scene reference point to the candidate camera and the closest selected camera. This describes whether the candidate observes the scene from a different position around it. Cameras located at different distances along the same ray from the reference point have zero angular separation, even if their positions are far apart.
\end{itemize}

\textbf{Image quality.}
The feature $\kappa_{\mathrm{sharp}}$ measures the sharpness of the candidate image. It provides information about image blur, allowing the selector to account for the quality of the image in addition to its camera position, viewing direction and appearance relationships.

The features are computed from the captured images and camera poses, using one DINO embedding per frame. Comparisons with the optimization views and image sharpness do not depend on the current selection, whereas the four features involving the selected frames are updated at every selection step. Feature computation and sequential scoring require no rendering, reconstruction or per-scene optimization.

\textbf{Additional feature groups tested.}
Besides removing features from the representation above, we tested six groups of \emph{additional}
features. These groups are evaluated in alternative feature configurations using the same training recipe; \autoref{tab:sppfeaturesweep} specifies the retained features and the additional group used in each configuration.
\begin{itemize}\setlength{\itemsep}{1pt}
  \item \emph{Worst-case terms} (3): the largest distance, the largest viewing-direction angle and the
  lowest DINO similarity to any optimization view. These are the pessimistic counterparts of
  $d_{\min}^{\gO}$, $\gamma_{\min}^{\gO}$ and $\rho_{\max}^{\gO}$, and describe
  the optimization view the candidate serves \emph{worst} rather than best.

  \item \emph{Capture context} (3): three quantities that are constant within a scene, namely the radius
  of the capture, the logarithm of the candidate-pool size, and the budget fraction $K/N$. They are
  identical for every candidate and let the network condition its scoring on the size and the scale of
  the capture rather than on the candidate alone.

  \item \emph{Rank-normalized} (3): the same three quantities $d_{\min}^{\gO}$,
  $d_{\mathrm{mean}}^{\gO}$ and $\gamma_{\mathrm{mean}}^{\gO}$, encoded differently.
  The three quantities are instead replaced by the position of the candidate in the sorted order, mapped linearly to $[-0.5, 0.5]$: the candidate nearest
  to the optimization views receives $-0.5$, the farthest $+0.5$, and the rest are spaced evenly in
  between. The encoding then depends only on the ordering of the candidates, not on the distances
  themselves, which also discards how large the gaps between them are.

  \item \emph{Soft aggregates} (3): smooth counterparts of $d_{\min}^{\gO}$,
  $\gamma_{\min}^{\gO}$ and $\rho_{\max}^{\gO}$, obtained by a temperature-weighted
  average over the optimization, so that a single very close
  optimization view cannot determine the value on its own.

  \item \emph{Mean redundancy} (3): redundancy with respect to the selected frames measured by the
  average over all selected frames rather than by the nearest one, for camera position, viewing
  direction and appearance. They complement $d_{\min}^{S}$, $\gamma_{\min}^{S}$
  and $\rho_{\max}^{S}$, which only look at the closest selected frame.

  \item \emph{Set-aware trio} (3): the fraction of optimization views for which the candidate would
  become the closest observed frame if it were selected, the marginal log-determinant gain of a
  determinantal kernel built from appearance similarity, and the distance to the second-nearest
  selected camera.
\end{itemize}
None of these groups improves on the deployed representation. The closest is the worst-case group at
$-0.009$\,dB, followed by capture context ($-0.011$), rank-normalized ($-0.015$), soft aggregates
($-0.041$) and mean redundancy ($-0.052$); the set-aware trio is the only clearly harmful one
($-0.295$\,dB). Together with the removals in ~\autoref{tab:sppfeaturesweep}, this indicates that the
representation is on a plateau: the quantities that matter are already present, and none of the tested alternative configurations outperforms the deployed 18-feature representation.

\begin{table}[t]
\centering
\scriptsize
\caption{\textbf{Tested features.} Each row is one feature configuration that does not improve on the deployed one: a mark means the feature is available to the network, a blank means its column is zeroed (width and parameter count are unchanged). $\Delta$PSNR is measured at $K{=}30$ on 30 ScanNet++ validation scenes, disjoint from the ScanNet++ scenes used in the main results, averaged over three training seeds, relative to the deployed configuration (first row, all 18 features). Negative values indicate lower PSNR than the final configuration. The extra column names the additional feature group used in each configuration, as described in the text.}
\label{tab:sppfeaturesweep}
\setlength{\tabcolsep}{2.2pt}
\resizebox{\linewidth}{!}{%
\begin{tabular}{rcccccccccccccccccc|l|r}
\toprule
\# & $d_{\min}^{\gO}$ & $d_{\mathrm{mean}}^{\gO}$ & $d_{25}^{\gO}$ & $\gamma_{\min}^{\gO}$ & $\gamma_{\mathrm{mean}}^{\gO}$ & $\rho_{\max}^{\gO}$ & $\rho_{\mathrm{mean}}^{\gO}$ & $\sigma_{\rho}$ & $\tau_{\mathrm{tri}}$ & $\alpha_{\mathrm{mean}}$ & $b_{\perp}$ & $b_{\parallel}$ & $c_{\mathrm{vis}}$ & $d_{\min}^{S}$ & $\gamma_{\min}^{S}$ & $\rho_{\max}^{S}$ & $\phi_{\mathrm{nov}}$ & $\kappa_{\mathrm{sharp}}$ & extra & $\Delta$PSNR \\
\cmidrule(lr){2-14}\cmidrule(lr){15-18}\cmidrule(lr){19-19}
 & \multicolumn{13}{c}{relation to the optimization views} & \multicolumn{4}{c}{relation to the selected frames} & \multicolumn{1}{c}{quality} & & \\
\midrule
\textbf{1} & \textbf{$\bullet$} & \textbf{$\bullet$} & \textbf{$\bullet$} & \textbf{$\bullet$} & \textbf{$\bullet$} & \textbf{$\bullet$} & \textbf{$\bullet$} & \textbf{$\bullet$} & \textbf{$\bullet$} & \textbf{$\bullet$} & \textbf{$\bullet$} & \textbf{$\bullet$} & \textbf{$\bullet$} & \textbf{$\bullet$} & \textbf{$\bullet$} & \textbf{$\bullet$} & \textbf{$\bullet$} & \textbf{$\bullet$} &  & \textbf{$0.000$} \\
2 & $\bullet$ & $\bullet$ & $\bullet$ & $\bullet$ & $\bullet$ & $\bullet$ & $\bullet$ & $\bullet$ &  &  &  &  &  & $\bullet$ & $\bullet$ & $\bullet$ & $\bullet$ & $\bullet$ & worst-case terms & $-0.009$ \\
3 & $\bullet$ & $\bullet$ & $\bullet$ & $\bullet$ & $\bullet$ & $\bullet$ & $\bullet$ & $\bullet$ &  &  &  &  &  & $\bullet$ & $\bullet$ & $\bullet$ & $\bullet$ & $\bullet$ & capture context & $-0.011$ \\
4 & $\bullet$ & $\bullet$ &  & $\bullet$ & $\bullet$ & $\bullet$ & $\bullet$ &  &  &  &  &  &  & $\bullet$ &  & $\bullet$ &  & $\bullet$ &  & $-0.012$ \\
5 & $\bullet$ & $\bullet$ & $\bullet$ & $\bullet$ & $\bullet$ & $\bullet$ & $\bullet$ & $\bullet$ &  &  &  &  &  & $\bullet$ & $\bullet$ & $\bullet$ & $\bullet$ & $\bullet$ &  & $-0.013$ \\
6 & $\bullet$ & $\bullet$ & $\bullet$ & $\bullet$ & $\bullet$ & $\bullet$ & $\bullet$ & $\bullet$ &  &  &  &  &  & $\bullet$ & $\bullet$ & $\bullet$ & $\bullet$ & $\bullet$ & rank-normalized & $-0.015$ \\
7 & $\bullet$ &  & $\bullet$ & $\bullet$ & $\bullet$ & $\bullet$ & $\bullet$ & $\bullet$ &  &  &  &  &  & $\bullet$ & $\bullet$ & $\bullet$ & $\bullet$ & $\bullet$ &  & $-0.016$ \\
8 & $\bullet$ & $\bullet$ & $\bullet$ & $\bullet$ & $\bullet$ & $\bullet$ & $\bullet$ & $\bullet$ &  &  &  &  &  & $\bullet$ & $\bullet$ & $\bullet$ & $\bullet$ &  &  & $-0.018$ \\
9 & $\bullet$ & $\bullet$ &  & $\bullet$ & $\bullet$ & $\bullet$ & $\bullet$ &  &  &  &  &  &  &  &  & $\bullet$ &  & $\bullet$ &  & $-0.031$ \\
10 & $\bullet$ & $\bullet$ & $\bullet$ & $\bullet$ & $\bullet$ & $\bullet$ & $\bullet$ & $\bullet$ &  &  &  &  &  & $\bullet$ & $\bullet$ & $\bullet$ & $\bullet$ & $\bullet$ & soft aggregates & $-0.041$ \\
11 &  &  &  &  &  & $\bullet$ & $\bullet$ &  &  &  &  &  &  &  &  & $\bullet$ &  &  &  & $-0.052$ \\
12 & $\bullet$ & $\bullet$ & $\bullet$ & $\bullet$ & $\bullet$ & $\bullet$ & $\bullet$ & $\bullet$ &  &  &  &  &  & $\bullet$ & $\bullet$ & $\bullet$ & $\bullet$ & $\bullet$ & mean redundancy & $-0.052$ \\
13 & $\bullet$ & $\bullet$ &  & $\bullet$ & $\bullet$ & $\bullet$ & $\bullet$ &  &  &  &  &  &  & $\bullet$ & $\bullet$ & $\bullet$ &  & $\bullet$ &  & $-0.061$ \\
14 & $\bullet$ & $\bullet$ & $\bullet$ & $\bullet$ & $\bullet$ &  &  & $\bullet$ &  &  &  &  &  & $\bullet$ & $\bullet$ &  & $\bullet$ & $\bullet$ &  & $-0.063$ \\
15 & $\bullet$ & $\bullet$ & $\bullet$ & $\bullet$ & $\bullet$ &  &  & $\bullet$ &  &  &  &  &  & $\bullet$ & $\bullet$ &  & $\bullet$ &  &  & $-0.067$ \\
16 & $\bullet$ & $\bullet$ & $\bullet$ &  &  & $\bullet$ & $\bullet$ & $\bullet$ &  &  &  &  &  & $\bullet$ &  & $\bullet$ & $\bullet$ &  &  & $-0.067$ \\
17 & $\bullet$ & $\bullet$ &  &  & $\bullet$ & $\bullet$ & $\bullet$ &  &  &  &  &  &  & $\bullet$ & $\bullet$ & $\bullet$ &  & $\bullet$ &  & $-0.078$ \\
18 & $\bullet$ &  &  &  &  &  &  &  &  &  &  &  &  & $\bullet$ &  & $\bullet$ &  &  &  & $-0.079$ \\
19 & $\bullet$ & $\bullet$ &  & $\bullet$ &  & $\bullet$ & $\bullet$ &  &  &  &  &  &  & $\bullet$ & $\bullet$ & $\bullet$ &  & $\bullet$ &  & $-0.083$ \\
20 & $\bullet$ & $\bullet$ &  & $\bullet$ & $\bullet$ & $\bullet$ & $\bullet$ &  &  &  &  &  &  & $\bullet$ & $\bullet$ & $\bullet$ &  &  &  & $-0.088$ \\
21 & $\bullet$ & $\bullet$ &  & $\bullet$ & $\bullet$ & $\bullet$ &  &  &  &  &  &  &  & $\bullet$ & $\bullet$ & $\bullet$ &  & $\bullet$ &  & $-0.094$ \\
22 & $\bullet$ & $\bullet$ &  &  &  & $\bullet$ & $\bullet$ &  &  &  &  &  &  & $\bullet$ & $\bullet$ & $\bullet$ &  & $\bullet$ &  & $-0.115$ \\
23 & $\bullet$ & $\bullet$ & $\bullet$ & $\bullet$ & $\bullet$ &  &  & $\bullet$ & $\bullet$ & $\bullet$ & $\bullet$ & $\bullet$ & $\bullet$ & $\bullet$ & $\bullet$ &  & $\bullet$ & $\bullet$ &  & $-0.124$ \\
24 &  & $\bullet$ &  & $\bullet$ & $\bullet$ & $\bullet$ & $\bullet$ &  &  &  &  &  &  & $\bullet$ & $\bullet$ & $\bullet$ &  & $\bullet$ &  & $-0.131$ \\
25 & $\bullet$ & $\bullet$ & $\bullet$ &  &  & $\bullet$ & $\bullet$ & $\bullet$ &  &  &  &  &  & $\bullet$ &  & $\bullet$ & $\bullet$ & $\bullet$ &  & $-0.132$ \\
26 & $\bullet$ & $\bullet$ &  & $\bullet$ & $\bullet$ & $\bullet$ & $\bullet$ &  &  &  &  &  &  & $\bullet$ & $\bullet$ &  &  & $\bullet$ &  & $-0.133$ \\
27 & $\bullet$ & $\bullet$ &  & $\bullet$ & $\bullet$ &  &  &  &  &  &  &  &  & $\bullet$ & $\bullet$ & $\bullet$ &  &  &  & $-0.148$ \\
28 & $\bullet$ & $\bullet$ &  & $\bullet$ & $\bullet$ & $\bullet$ & $\bullet$ &  &  &  &  &  &  & $\bullet$ & $\bullet$ &  &  &  &  & $-0.151$ \\
29 & $\bullet$ & $\bullet$ &  & $\bullet$ & $\bullet$ &  &  &  &  &  &  &  &  & $\bullet$ & $\bullet$ & $\bullet$ &  & $\bullet$ &  & $-0.163$ \\
30 & $\bullet$ & $\bullet$ &  & $\bullet$ & $\bullet$ &  &  &  &  &  &  &  &  & $\bullet$ & $\bullet$ &  &  & $\bullet$ &  & $-0.167$ \\
31 & $\bullet$ & $\bullet$ &  & $\bullet$ & $\bullet$ &  & $\bullet$ &  &  &  &  &  &  & $\bullet$ & $\bullet$ & $\bullet$ &  & $\bullet$ &  & $-0.182$ \\
32 & $\bullet$ & $\bullet$ & $\bullet$ &  &  &  &  & $\bullet$ &  &  &  &  &  & $\bullet$ &  & $\bullet$ & $\bullet$ &  &  & $-0.187$ \\
33 & $\bullet$ & $\bullet$ &  & $\bullet$ & $\bullet$ &  &  &  &  &  &  &  &  & $\bullet$ & $\bullet$ &  &  &  &  & $-0.197$ \\
34 & $\bullet$ & $\bullet$ &  &  &  &  &  &  &  &  &  &  &  & $\bullet$ &  &  &  &  &  & $-0.252$ \\
35 & $\bullet$ & $\bullet$ & $\bullet$ & $\bullet$ & $\bullet$ & $\bullet$ & $\bullet$ & $\bullet$ &  &  &  &  &  & $\bullet$ & $\bullet$ & $\bullet$ & $\bullet$ & $\bullet$ & set-aware trio & $-0.295$ \\
36 &  &  &  & $\bullet$ & $\bullet$ &  &  &  &  &  &  &  &  &  & $\bullet$ &  &  &  &  & $-0.717$ \\
37 & $\bullet$ &  &  &  &  &  &  &  &  &  &  &  &  &  &  &  &  &  &  & $-1.624$ \\
38 & $\bullet$ & $\bullet$ &  & $\bullet$ & $\bullet$ & $\bullet$ & $\bullet$ &  &  &  &  &  &  &  &  &  &  &  &  & $-1.869$ \\
39 & $\bullet$ & $\bullet$ &  & $\bullet$ & $\bullet$ & $\bullet$ & $\bullet$ &  &  &  &  &  &  &  &  &  &  & $\bullet$ &  & $-2.048$ \\
40 & $\bullet$ & $\bullet$ &  &  &  &  &  &  &  &  &  &  &  &  &  &  &  &  &  & $-2.543$ \\
41 &  &  &  &  &  &  &  &  &  &  &  &  &  & $\bullet$ & $\bullet$ & $\bullet$ &  &  &  & $-4.483$ \\
42 &  &  &  &  &  &  &  &  &  &  &  &  &  & $\bullet$ & $\bullet$ & $\bullet$ &  & $\bullet$ &  & $-6.610$ \\
43 &  & $\bullet$ &  &  & $\bullet$ &  & $\bullet$ &  &  &  &  &  &  &  &  &  &  & $\bullet$ &  & $-7.611$ \\
\bottomrule
\end{tabular}}
\end{table}

\subsection{Genetic-search supervision}
\label{app:ga_details}

\textbf{Population Initialization.} Each individual is a sorted set of $K$ distinct indices from the candidate pool. We use a population
of $C{=}40$ individuals. The initial population contains $16$ spatial farthest-view subsets from different starting anchors within the target region defined by coverage level $p$, $12$ uniformly spaced subsets with
different phase offsets along the region's trajectory span, and $12$ subsets drawn at random from
the region.

\textbf{Fitness Evaluation.} Each subset is passed to the reconstructor, and the optimization views are rendered to obtain PSNR, SSIM, and LPIPS. The three metrics are independently min-max normalized over all distinct subsets evaluated for that scene, with LPIPS negated, and then averaged with equal weights. Reconstruction results are cached using the sorted subset indices, so a repeated individual is not rendered again.

\textbf{Selection, Crossover, and Mutation.} The two highest-fitness individuals are copied unchanged to the next generation; the remaining
$38$ slots are filled with children, each produced independently as follows. First, two parents are
selected by tournament: for each parent, four individuals are drawn uniformly at random from the
population and the fittest of the four is kept. Selection is therefore stochastic but biased toward
high fitness: any individual can enter a tournament, but only the strongest entrant becomes a
parent. Second, with probability $0.8$ the child is created by crossover, drawing $K$ distinct
frames from the union of the two parents; otherwise it is a copy of the first parent. Third,
independently of whether crossover occurred, mutation is applied with probability $0.3$ and replaces
$\max(2,\lfloor K/5\rfloor)$ of the child's frames with randomly drawn candidate frames; duplicates are removed and refilled at random, keeping exactly $K$ distinct views. Crossover
and mutation are thus not exclusive: a child can undergo both, either one alone, or neither, in
which case it is an unmodified copy of a tournament winner.

\textbf{Termination and Retained Solutions.}
The search runs for $G=15$ generations. We store the 30 highest-fitness subsets and use the best $M=4$ as training trajectories. All training labels are generated at a single budget, $K{=}20$, yet the selector generalizes across
budgets: it is evaluated at every $K$ from $10$ to $50$ and outperforms the strongest baselines (\autoref{fig:heatmap}), since the budget $K$ 
enters only as the number of greedy steps taken at inference. No learned model is used to initialize or guide the search. We also explored automated hyperparameter optimization, but it produced no meaningful improvement over the fixed configuration.

\subsection{Training details}
\label{app:training_details}

\textbf{Training Data and Splits.}
The selector used throughout the paper is trained on DL3DV alone. Of the $141$ labeled scenes, $21$ are held out as the DL3DV test set, $20$ as a validation set, and a
further $30$ are excluded because they were used to tune and validate the genetic search that
produces the labels, leaving $70$ training scenes. Each is relabeled at five \emph{spread} levels, which vary
how wide a target region the request covers and hence how many optimization views represent it,
$4$, $6$, $9$, $12$ and $15$ views respectively. A scene at a given spread level is its own training
item. During
label generation the number of target views scales with the level, $\max(4,\lfloor 0.15p \rceil)$,
giving $4$, $6$, $9$, $12$ and $15$ targets, and each target is assigned one optimization view. Evaluation uses a fixed target count (15 frames). Training scenes are disjoint from the scenes used to tune the genetic search and from all evaluation splits. This is the only data the model sees: ScanNet++,
ScanNet-iPhone, 7-Scenes, Tanks and Temples, Mip-NeRF~360 and CO3D are all evaluated zero-shot, with
no fine-tuning and no labels of their own.

\textbf{Supervision from Genetic Search.}
Labels come from the genetic search described in Appendix~\ref{app:ga_details}, run at $K{=}20$. Of the $30$
highest-fitness subsets it retains per item, the top $4$ are used for training, weighted by
$\exp\!\big((f-f_{\max})/\sigma_f\big)$ normalized to unit mean so the loss scale does not depend on
how many subsets are kept. Each subset is converted into $K$ per-step targets by the ordering of
\autoref{eq:order}, and the model is trained by autoregressive per-step cross-entropy: at step $t$
the scorer emits one logit per candidate, already-selected frames are masked out, and the target is
the frame the ordering places at that position. The loss is averaged over the $K$ steps of a subset
and one optimizer step is taken per (item, subset) pair, giving $350 \times 4 \times 10 = 14{,}000$
updates over $10$ epochs.

\textbf{Network and Optimization.}
The network is the three-layer MLP of width $64$ described above: $18$ features (a 19th input column is held at zero throughout), GELU activations~\cite{hendrycks2016gelu},
$5{,}505$ parameters. We use Adam~\cite{kingma2015adam} at a fixed learning rate of $10^{-3}$. All reported numbers are averaged over three training seeds per scene, which differ
in initialization and data-shuffling order only. Evaluation additionally averages over three target draws per scene.

\textbf{Seed variability.}
The spread across seeds is small. Over all six datasets and all five budgets, the standard deviation
of the dataset mean across the three training seeds stays below the margins we
report over the strongest baseline (Table~\ref{tab:seedvar}). Thus, our conclusions do not depend on which seed is used. 
All methods are evaluated on identical scenes and held-out target views, ensuring a paired comparison.

\begin{table}[t]
\centering
\small
\setlength{\tabcolsep}{6pt}
\renewcommand{\arraystretch}{1.05}
\caption{\textbf{Seed variability.} Standard deviation (dB) of the dataset-mean PSNR across the three
training seeds, which differ only in initialization and in the order training items are presented.
$n$ is the number of evaluation scenes.}
\label{tab:seedvar}
\begin{tabular}{l c ccccc}
\toprule
\multirow{2}{*}{Dataset} & \multirow{2}{*}{$n$} & \multicolumn{5}{c}{Budget $K$} \\
\cmidrule(lr){3-7}
 & & $10$ & $20$ & $30$ & $40$ & $50$ \\
\midrule
DL3DV            & 21 & 0.022 & 0.011 & 0.043 & 0.013 & 0.012 \\
Tanks \& Temples &  6 & 0.089 & 0.030 & 0.070 & 0.035 & 0.139 \\
Mip-NeRF\,360    &  9 & 0.049 & 0.136 & 0.053 & 0.031 & 0.043 \\
ScanNet-iPhone   & 30 & 0.004 & 0.103 & 0.080 & 0.055 & 0.032 \\
ScanNet++        & 30 & 0.059 & 0.069 & 0.028 & 0.074 & 0.052 \\
7-Scenes         &  7 & 0.097 & 0.040 & 0.073 & 0.105 & 0.085 \\
\bottomrule
\end{tabular}
\end{table}

\textbf{Scalability to full captures.} The natural alternative would be to let a network read the capture directly (all candidate poses, and possibly all candidate images), and predict a subset. We deliberately avoid this, because the pool is an entire capture rather than a small number of views. In our benchmarks, a single scene offers from a few hundred to several thousand candidate frames (up to 2,781 on ScanNet++, and more than 10{,}000 on 7-Scenes), whereas feed-forward reconstructors are trained at a fixed and much smaller number of input views, and attention over candidates costs $\mathcal{O}(n^2)$ in the pool size. A selector built that way would defeat the purpose of a selector designed for fast inference on any incoming video.

Our representation avoids this issue. Each candidate is reduced to a fixed-length vector by aggregating over the optimization views and over the selected set. Thus, width of the input is independent of how many frames the capture contains, of how many optimization views are requested, and of how many frames have already been selected. Cost is linear in the pool size, the same network applies unchanged to a 300-frame and a 10,000-frame capture, and nothing about the architecture encodes capture length or capture order.

\subsection{Reconstructor details}
\label{app:reconstructor}

\textbf{Choice of reconstructor.}
We use LongLRM as feed-forward reconstructor throughout the main experiments. On the same 21 held-out DL3DV scenes, it is between $6$ and $8$\,dB ahead of AnySplat and Wild3R at every budget we study (Table~\ref{tab:reconchoice}), and it is
the only one of the three that can consume the full capture: feeding all ${\sim}130$ frames exhausts a
$40$\,GB A100 GPU for AnySplat and Wild3R on all $21$ scenes, whereas LongLRM completes them at $17.8$\,GB
peak memory. AnySplat and Wild3R are designed for fewer input views, which is also visible in how
little they gain between $K{=}20$ and $K{=}40$ ($+0.38$ and $+0.42$\,dB, against $+1.71$\,dB for
LongLRM).

\begin{table}[t]
\centering
\small
\caption{\textbf{Comparison across reconstructors.} Novel view synthesis on the 21 DL3DV scenes with
the deployed selector (three training seeds), for three feed-forward reconstructors. Each cell is
PSNR$\uparrow$ / SSIM$\uparrow$ / LPIPS$\downarrow$. The last row feeds all candidate frames
(${\sim}130$ on average) instead of a selected subset.}
\label{tab:reconchoice}
\begin{tabular}{lccc}
\toprule
Input & LongLRM & AnySplat & Wild3R \\
\midrule
$K=20$            & $20.97 / 0.696 / 0.204$ & $14.62 / 0.381 / 0.398$ & $14.99 / 0.414 / 0.387$ \\
$K=30$            & $22.33 / 0.724 / 0.177$ & $14.88 / 0.390 / 0.390$ & $15.27 / 0.423 / 0.383$ \\
$K=40$            & $22.68 / 0.734 / 0.171$ & $15.00 / 0.393 / 0.389$ & $15.41 / 0.428 / 0.382$ \\
\midrule

all frames (${\sim}130$) & $21.00 / 0.706 / 0.205$ & out of memory & out of memory \\
\bottomrule
\end{tabular}
\end{table}

\section{Additional Ablations}
\label{app:ablations}

\textbf{Scorer architecture ablation.}
We evaluate whether more expressive scorer architectures improve over our architecture. The full comparison in \autoref{tab:scorer_arch} covers wider and deeper MLPs, candidate self-attention, learned attention over optimization views,
recurrent decoders (LSTM~\cite{hochreiter1997long}, GRU~\cite{chung2014empirical}) and selected-set conditioning. None improves
beyond the resolution of the experiment: the largest positive difference is $+0.004$\,dB.


\begin{table}[t]
\centering
\caption{Scorer-architecture comparison on 21 held-out DL3DV scenes
($K=20$).We report PSNR differences relative to our MLP, which was fixed in advance:
no architecture was selected on the basis of these numbers.}
\label{tab:scorer_arch}
\small
\begin{tabular}{lc}
\toprule
Scorer & $\Delta$PSNR (dB) $\uparrow$ \\
\midrule
Wider MLP (256 units)                            & $+0.004$ \\
3-layer MLP (ours)                               & $0.000$ \\
Deep residual MLP                                & $-0.007$ \\
MLP with pooled candidate-set context            & $-0.015$ \\
Cross-attention to selected frames               & $-0.027$ \\
Candidate self-attention (1 layer)               & $-0.095$ \\
Self-attention with capture-order encoding       & $-0.119$ \\
Candidate self-attention (2 layers)              & $-0.136$ \\
Iterative candidate rescoring                    & $-0.167$ \\
Self-attention with an LSTM decoder              & $-0.218$ \\
Bidirectional LSTM over capture order            & $-0.282$ \\
LSTM-based sequential scorer                     & $-0.313$ \\
Temporal convolution over capture order          & $-0.330$ \\
LSTM scorer with 128 hidden units                & $-0.343$ \\
GRU-based sequential scorer                      & $-0.349$ \\
RNN-based sequential scorer                      & $-0.391$ \\
Attention over optimization views                & $-0.551$ \\
\bottomrule
\end{tabular}
\end{table}

\textbf{Comparison of feature configurations.}
We compare our 18-feature configuration described above with alternatives that add, remove or replace groups of features. All variants use the same network architecture and training procedure. 

Table~\ref{tab:sppfeaturesweep} reports the mean PSNR difference between some of the alternatives explored and the final configuration under the standard protocol with $K{=}30$. Each configuration is trained with three seeds. 

The final configuration achieves the highest mean PSNR among the tested representations. The results support the choice of the final representation.

\subsection{Comparison with the Genetic Search}
\label{app:ga_vs_learned}

The selector is trained to imitate the genetic search, raising the question of how much performance is lost through imitation. We evaluate this on the $21$ held-out DL3DV scenes at $K{=}20$, the budget used for label generation. For each scene, we evaluate every subset reconstructed during the search on the held-out target views. These subsets achieve an average PSNR of $20.07$\,dB, while the best-selected subset generated with GA achieves $20.99$\,dB. Under an identical rendering budget, uniform random search achieves $20.37$\,dB when the best subset per scene is selected using the same fitness criterion. The genetic search therefore improves over random search by $0.62$\,dB without additional reconstruction cost. Our selector achieves $20.97$\,dB, within $0.02$\,dB of the search it imitates and outperforming it on $12$ of the $21$ scenes, while requiring only a single forward pass per selection step and no reconstruction at inference.

\paragraph{Contribution of learned feature aggregation.}
Our selector uses target-conditioned features unavailable to most baselines, raising the question of whether its advantage stems from learning or from access to these signals alone. To isolate the contribution of learning, we evaluate an untrained control with the same feature representation and greedy selection procedure. The control computes the same $18$ features, standardizes each across candidates using z-scores, and sums them after reversing the sign of features for which lower values are preferable, such as target distance and redundancy. It greedily selects the highest-scoring frame and updates the selection-dependent features until $K$ frames are selected, without learned weights or a network. The selected frames are passed to LongLRM, and PSNR is evaluated on the same target views. As shown in Table~\ref{tab:handsign}, the learned scorer outperforms this control in $29$ of the $30$ dataset-budget combinations, achieving an average PSNR improvement of $0.48$\,dB. These results indicate that, even with identical input information and selection procedures, learning to combine the features provides a consistent advantage over untrained aggregation.

\begin{table}[t]
\centering
\small
\setlength{\tabcolsep}{6pt}
\renewcommand{\arraystretch}{1.05}
\caption{\textbf{Learned versus untrained feature aggregation.} PSNR difference (dB) between our learned selector and a
control that uses the selector's own representation and greedy loop with no learned scorer, ranking
candidates by the sum of their sign-corrected features with nothing fitted. Positive values favour
the learned scorer, which is ahead in $29$ of the $30$ cells, by $0.48$\,dB on average. $n$ is the
number of evaluation scenes.}
\label{tab:handsign}
\begin{tabular}{l c ccccc}
\toprule
\multirow{2}{*}{Dataset} & \multirow{2}{*}{$n$} & \multicolumn{5}{c}{Budget $K$} \\
\cmidrule(lr){3-7}
 & & $10$ & $20$ & $30$ & $40$ & $50$ \\
\midrule
DL3DV            & 21 & $+0.17$ & $+0.51$ & $+0.29$ & $+0.15$ & $+0.17$ \\
Tanks \& Temples &  6 & $-0.40$ & $+0.02$ & $+0.18$ & $+0.30$ & $+0.47$ \\
Mip-NeRF\,360    &  9 & $+0.45$ & $+0.41$ & $+0.63$ & $+0.72$ & $+0.60$ \\
ScanNet-iPhone   & 30 & $+0.31$ & $+0.57$ & $+0.62$ & $+0.54$ & $+0.52$ \\
ScanNet++        & 30 & $+0.42$ & $+0.87$ & $+0.57$ & $+0.49$ & $+0.46$ \\
7-Scenes         &  7 & $+0.61$ & $+0.94$ & $+0.89$ & $+0.95$ & $+0.97$ \\
\bottomrule
\end{tabular}
\end{table}

\section{Additional Results}
\label{app:qualitative_results}

\subsection{Performance across input budgets}
\label{app:budget_sweep}

Table~\ref{tab:ksweep_all} reports the full quantitative results underlying the summary heatmap in \autoref{fig:heatmap}, across all datasets and budgets $K\in{10,20,30,40,50}$. Our selector achieves higher PSNR than the strongest external baseline in all dataset-budget combinations, confirming that the gains shown in the heatmap are consistent across budgets and datasets.

\begin{table*}[t]
\centering
\caption{\textbf{Input budget sweep across datasets}.
Each cell reports PSNR$\uparrow$~/~SSIM$\uparrow$~/~LPIPS$\downarrow$.
Baselines: FVS and NeRF-Director~\cite{xiao2024nerfdirector}, the MVSNet
view-selection score~\cite{yao2018mvsnet}, DPP~\cite{kulesza2012dpp},
$k$-medoids~\cite{kaufman2009finding}, FisherRF~\cite{jiang2024fisherrf},
and COVER~\cite{chen2026cover}. We report our selector with and without appearance features.}

\label{tab:ksweep_all}

\tiny
\setlength{\tabcolsep}{1.5pt}

\resizebox{\textwidth}{!}{%
\begin{tabular}{lccccc|ccccc}
\toprule
& \multicolumn{5}{c}{\textbf{DL3DV} ($n{=}21$)}
& \multicolumn{5}{c}{\textbf{Tanks and Temples} ($n{=}6$)} \\
\cmidrule(lr){2-6}
\cmidrule(lr){7-11}

Selection
& $K{=}10$ & $K{=}20$ & $K{=}30$ & $K{=}40$ & $K{=}50$ & $K{=}10$ & $K{=}20$ & $K{=}30$ & $K{=}40$ & $K{=}50$ \\
\midrule

Random
& 14.58/.526/.424 & 19.63/.647/.255 & 21.14/.686/.213 & 21.72/.701/.197 & 22.09/.717/.184
& 11.64/.419/.606 & 16.52/.538/.406 & 17.83/.580/.340 & 18.49/.599/.317 & 18.76/.606/.294 \\

Uniform
& 14.94/.531/.413 & 20.18/.660/.239 & 21.60/.696/.201 & 22.32/.719/.181 & 22.43/.727/.176
& 12.24/.427/.579 & 17.10/.549/.386 & 18.39/.591/.315 & 19.27/.616/.280 & 19.29/.617/.273 \\

FVS
& 15.21/.534/.405 & 20.61/.674/.224 & 21.99/.711/.187 & 22.51/.727/.174 & 22.74/.736/.168
& 12.15/.408/.620 & 17.05/.541/.399 & 18.30/.586/.330 & 18.82/.604/.297 & 19.04/.612/.287 \\

NeRF-Director
& 15.22/.531/.410 & 20.47/.672/.225 & 21.75/.701/.191 & 22.25/.716/.177 & 22.44/.725/.170
& 12.09/.406/.625 & 16.82/.534/.402 & 18.21/.583/.329 & 18.62/.600/.302 & 18.79/.603/.295 \\

MVSNet score
& 15.04/.555/.374 & 20.22/.675/.222 & 21.74/.713/.187 & 22.41/.731/.174 & 22.63/\textbf{.740}/.168
& 12.18/.436/.537 & 17.41/.575/.341 & 18.71/.618/.286 & 19.58/.644/.257 & 19.83/.652/.248 \\

DPP
& 14.94/.526/.409 & 20.27/.661/.235 & 21.59/.697/.197 & 22.19/.716/.180 & 22.47/.727/.172
& 11.63/.404/.631 & 16.06/.519/.450 & 17.80/.563/.358 & 18.40/.587/.320 & 18.76/.600/.297 \\

$k$-medoids
& 15.15/.533/.408 & 20.31/.661/.232 & 21.68/.699/.194 & 22.26/.718/.179 & 22.52/.727/.171
& 12.23/.417/.583 & 16.57/.533/.417 & 18.06/.580/.330 & 18.83/.604/.295 & 18.93/.606/.285 \\

FisherRF
& 14.75/.532/.404 & 20.09/.659/.242 & 21.61/.700/.200 & 22.28/.719/.182 & 22.44/.726/.177
& 11.49/.416/.607 & 16.43/.532/.408 & 18.17/.581/.342 & 18.67/.596/.315 & 19.08/.607/.304 \\

COVER
& 14.63/.529/.414 & 20.31/.661/.234 & 21.59/.706/.204 & 22.34/.724 /.188 & 22.39/.725/.176
& 11.89/.411/.603 & 16.66/.529/.422 & 17.99/.567/.344 & 18.41/.588/.319 & 18.84/.601/.301 \\

\textbf{Ours, no appearance}
& 15.69/\textbf{.572}/.349 & 20.85/.694/.208 & 22.30/\textbf{.725}/.178 & 22.66/\textbf{.734}/.171 & 22.72/.738/.168
& \textbf{13.16}/\textbf{.440}/\textbf{.530} & 18.04/.590/.323 & \textbf{19.47}/.633/.267 & 20.02/.649/.247 & 20.35/.658/.234 \\

\textbf{Ours}
& \textbf{15.69}/.571/\textbf{.349} & \textbf{20.97}/\textbf{.696}/\textbf{.204} & \textbf{22.33}/.724/\textbf{.177} & \textbf{22.68}/.734/\textbf{.171} & \textbf{22.76}/.738/\textbf{.168}
& 12.98/.433/.545 & \textbf{18.24}/\textbf{.595}/\textbf{.312} & 19.46/\textbf{.633}/\textbf{.263} & \textbf{20.10}/\textbf{.653}/\textbf{.241} & \textbf{20.37}/\textbf{.661}/\textbf{.231} \\

\bottomrule
\end{tabular}}
\vspace{6pt}

\resizebox{\textwidth}{!}{%
\begin{tabular}{lccccc|ccccc}
\toprule
& \multicolumn{5}{c}{\textbf{Mip-NeRF~360} ($n{=}9$)}
& \multicolumn{5}{c}{\textbf{ScanNet-iPhone} ($n{=}30$)} \\
\cmidrule(lr){2-6}
\cmidrule(lr){7-11}

Selection
& $K{=}10$ & $K{=}20$ & $K{=}30$ & $K{=}40$ & $K{=}50$ & $K{=}10$ & $K{=}20$ & $K{=}30$ & $K{=}40$ & $K{=}50$ \\
\midrule

Random
& 11.91/.319/.618 & 17.57/.423/.439 & 18.96/.458/.378 & 19.89/.481/.334 & 20.39/.497/.311
& 12.77/.589/.581 & 16.02/.634/.466 & 17.02/.648/.415 & 17.57/.657/.383 & 17.88/.661/.364 \\

Uniform
& 12.53/.325/.614 & 18.20/.434/.413 & 19.34/.465/.360 & 20.32/.489/.316 & 20.55/.502/.300
& 12.70/.588/.592 & 16.52/.642/.443 & 17.37/.654/.393 & 17.83/.661/.366 & 18.04/.663/.350 \\

FVS
& 13.02/.321/.631 & 18.55/.433/.433 & 19.79/.473/.353 & 20.36/.497/.314 & 20.52/.500/.301
& 12.75/.588/.594 & 16.12/.637/.467 & 17.05/.650/.410 & 17.46/.654/.381 & 17.72/.658/.363 \\

NeRF-Director
& 12.89/.316/.644 & 18.69/.433/.433 & 20.06/.476/.349 & 20.51/.495/.315 & 20.67/.502/.301
& 12.75/.588/.595 & 16.10/.637/.468 & 17.10/.650/.410 & 17.53/.656/.380 & 17.80/.659/.362 \\

MVSNet score
& 12.87/.339/.553 & 18.41/.462/.364 & 19.88/.503/.313 & 20.36/.520/.294 & 20.53/.525/.287
& 13.32/.600/.527 & 16.17/.637/.440 & 16.78/.645/.419 & 17.06/.648/.407 & 17.15/.648/.403 \\

DPP
& 11.71/.310/.643 & 18.42/.431/.437 & 19.85/.470/.368 & 20.35/.488/.333 & 20.58/.498/.315
& 12.56/.583/.597 & 16.17/.637/.471 & 17.15/.652/.410 & 17.54/.656/.382 & 17.82/.659/.363 \\

$k$-medoids
& 12.98/.325/.604 & 18.65/.438/.416 & 20.00/.477/.337 & 20.42/.491/.315 & 20.61/.497/.303
& 13.41/.597/.539 & 16.81/.647/.425 & 17.52/.657/.384 & 17.82/.661/.360 & 17.97/.662/.348 \\

FisherRF
& 11.95/.320/.622 & 16.53/.411/.466 & 18.50/.456/.386 & 19.32/.469/.351 & 19.70/.478/.340
& 11.84 /.606 /.589 & 15.96/.664 /.438 & 17.74/.694 /.365 & 18.35/.702/.337 & 18.99/.713/.306 \\

COVER~\cite{chen2026cover}
& 11.87/.323/.600 & 17.63/.421/.441 & 19.53/.467/.358 & 20.13/.483/.329 & 20.29/.487/.312
& 11.87/.606/.595 & 16.31/.666/.442 & 18.01/.693/.365 & 18.68/.707/.328 & 19.13/.713/.307 \\

\textbf{Ours, no appearance}
& 13.72/.351/.535 & 19.24/\textbf{.483}/.347 & 20.56/\textbf{.520}/.300 & 21.07/.534/.280 & 21.16/.535/\textbf{.273}
& 13.86/.638/.471 & 18.96/.718/.296 & 20.25/.739/.255 & 20.90/.748/.236 & 21.22/.754/.226 \\

\textbf{Ours}
& \textbf{13.78}/\textbf{.352}/\textbf{.534} & \textbf{19.27}/.481/\textbf{.346} & \textbf{20.63}/.520/\textbf{.297} & \textbf{21.08}/\textbf{.534}/\textbf{.277} & \textbf{21.17}/\textbf{.536}/.273
& \textbf{13.99}/\textbf{.639}/\textbf{.458} & \textbf{19.16}/\textbf{.722}/\textbf{.285} & \textbf{20.42}/\textbf{.743}/\textbf{.249} & \textbf{21.00}/\textbf{.752}/\textbf{.232} & \textbf{21.32}/\textbf{.757}/\textbf{.222} \\

\bottomrule
\end{tabular}}
\vspace{6pt}

\resizebox{\textwidth}{!}{%
\begin{tabular}{lccccc|ccccc}
\toprule
& \multicolumn{5}{c}{\textbf{ScanNet++} ($n{=}30$)}
& \multicolumn{5}{c}{\textbf{7-Scenes} ($n{=}7$)} \\
\cmidrule(lr){2-6}
\cmidrule(lr){7-11}

Selection
& $K{=}10$ & $K{=}20$ & $K{=}30$ & $K{=}40$ & $K{=}50$ & $K{=}10$ & $K{=}20$ & $K{=}30$ & $K{=}40$ & $K{=}50$ \\
\midrule

Random
& 15.39/.708/.381 & 20.77/.791/.228 & 23.04/.823/.177 & 24.11/.838/.156 & 24.58/.847/.146
& 14.60/.600/.520 & 18.35/.657/.360 & 19.22/.675/.317 & 19.68/.683/.296 & 19.74/.685/.288 \\

Uniform
& 15.77/.711/.372 & 21.68/.806/.202 & 23.72/.834/.162 & 24.78/.851/.141 & 25.17/.858/.131
& 15.19/.607/.481 & 18.39/.664/.339 & 19.20/.674/.317 & 19.79/.689/.294 & 20.00/.689/.282 \\

FVS
& 15.18/.685/.452 & 20.46/.771/.265 & 22.52/.808/.199 & 23.53/.826/.170 & 24.21/.839/.156
& 15.04/.616/.486 & 18.68/.670/.327 & 19.63/.683/.297 & 19.88/.688/.285 & 20.08/.691/.278 \\

NeRF-Director
& 14.83/.674/.471 & 19.99/.760/.285 & 21.99/.794/.217 & 22.80/.809/.194 & 23.29/.817/.181
& 14.87/.608/.488 & 18.92/.673/.322 & 19.80/.685/.291 & 20.14/.693/.278 & 20.28/.696/.271 \\

MVSNet score
& 15.94/.719/.356 & 20.68/.796/.219 & 22.26/.817/.190 & 23.18/.828/.173 & 23.47/.833/.167
& 13.99/.594/.527 & 17.93/.656/.379 & 18.92/.671/.347 & 19.00/.671/.336 & 19.20/.674/.328 \\

DPP
& 15.37/.687/.429 & 20.51/.773/.259 & 22.58/.811/.196 & 23.67/.829/.169 & 24.29/.840/.153
& 14.25/.589/.544 & 17.97/.646/.379 & 19.05/.664/.329 & 19.73/.679/.301 & 19.94/.685/.289 \\

$k$-medoids
& 16.46/.721/.348 & 21.65/.800/.207 & 23.29/.825/.170 & 24.13/.839/.151 & 24.59/.848/.143
& 15.13/.621/.463 & 19.11/.676/.325 & 19.80/.685/.288 & 20.18/.694/.273 & 20.31/.696/.270 \\

FisherRF
& 15.98/.720/.359 & 20.98/.787/.231 & 23.10/.823/.179 & 23.71/.828/.168 & 24.30/.840/.153
& 14.95/.605/.476 & 18.49/.662/.348 & 19.10/.669/.332 & 19.51/.680/.314 & 19.36/.676/.314 \\

COVER~\cite{chen2026cover}
& 16.08/.722/.351 & 21.20/.791/.230 & 22.77/.815/.189 & 23.52/.826/.174 & 23.96/.835/.163
& 14.53/.602/.490 & 19.00/.671/.337 & 19.81/.685/.304 & 20.14/.689/.283 & 20.02/.687/.284 \\

\textbf{Ours, no appearance}
& 17.20/.755/.284 & 23.02/.833/.162 & 24.54/.850/.142 & 25.24/.860/.132 & 25.58/.865/.126
& 16.03/.638/.408 & 20.24/\textbf{.701}/.280 & \textbf{20.83}/\textbf{.710}/\textbf{.257} & 20.96/.712/\textbf{.250} & 21.06/\textbf{.715}/\textbf{.246} \\

\textbf{Ours}
& \textbf{17.35}/\textbf{.758}/\textbf{.277} & \textbf{23.14}/\textbf{.836}/\textbf{.157} & \textbf{24.68}/\textbf{.853}/\textbf{.138} & \textbf{25.36}/\textbf{.861}/\textbf{.129} & \textbf{25.65}/\textbf{.867}/\textbf{.124}
& \textbf{16.15}/\textbf{.640}/\textbf{.406} & \textbf{20.32}/.700/\textbf{.276} & 20.80/.707/.260 & \textbf{21.04}/\textbf{.713}/.250 & \textbf{21.07}/.715/.246 \\

\bottomrule
\end{tabular}}
\vspace{6pt}

\end{table*}

\subsection{Further qualitative results}
\label{app:qualitative}
\begin{figure*}[t]
    \centering
    \includegraphics[width=\textwidth]{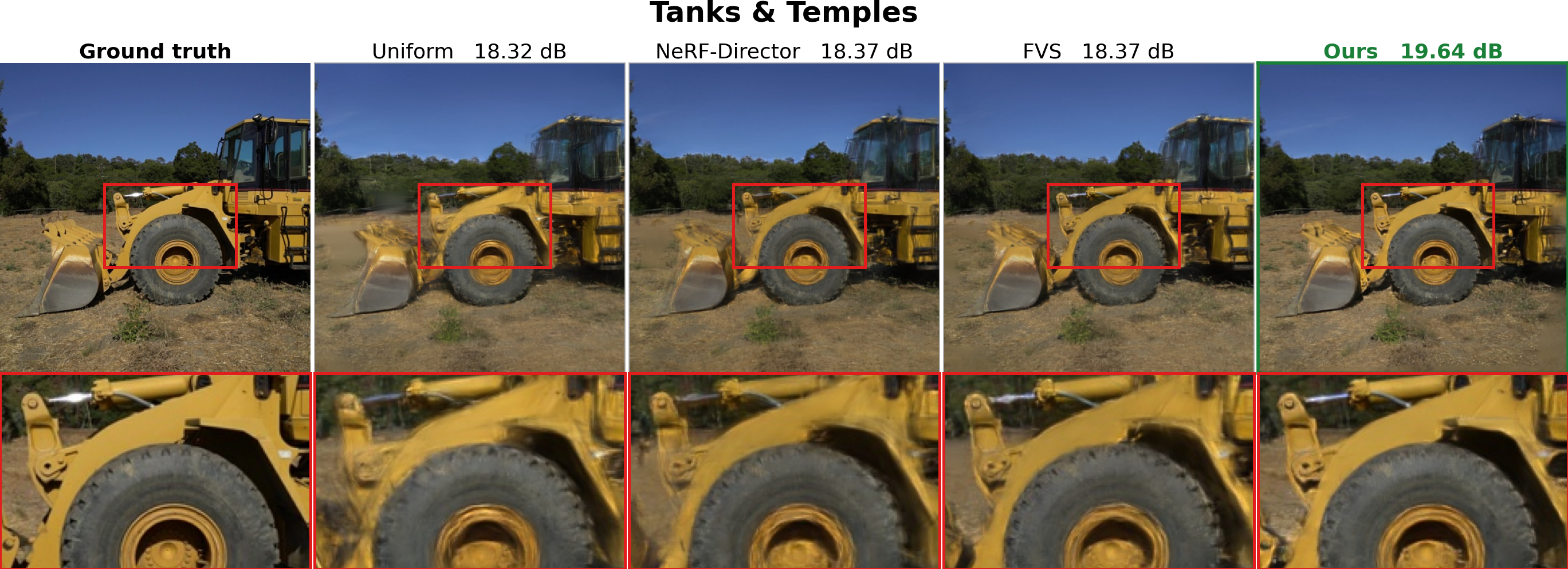}
    \caption{\textbf{Qualitative comparison on Tanks and Temples} at $K{=}40$.}
    \label{fig:qual_tandt}
\end{figure*}

\begin{figure*}[t]
    \centering
    \includegraphics[width=\textwidth]{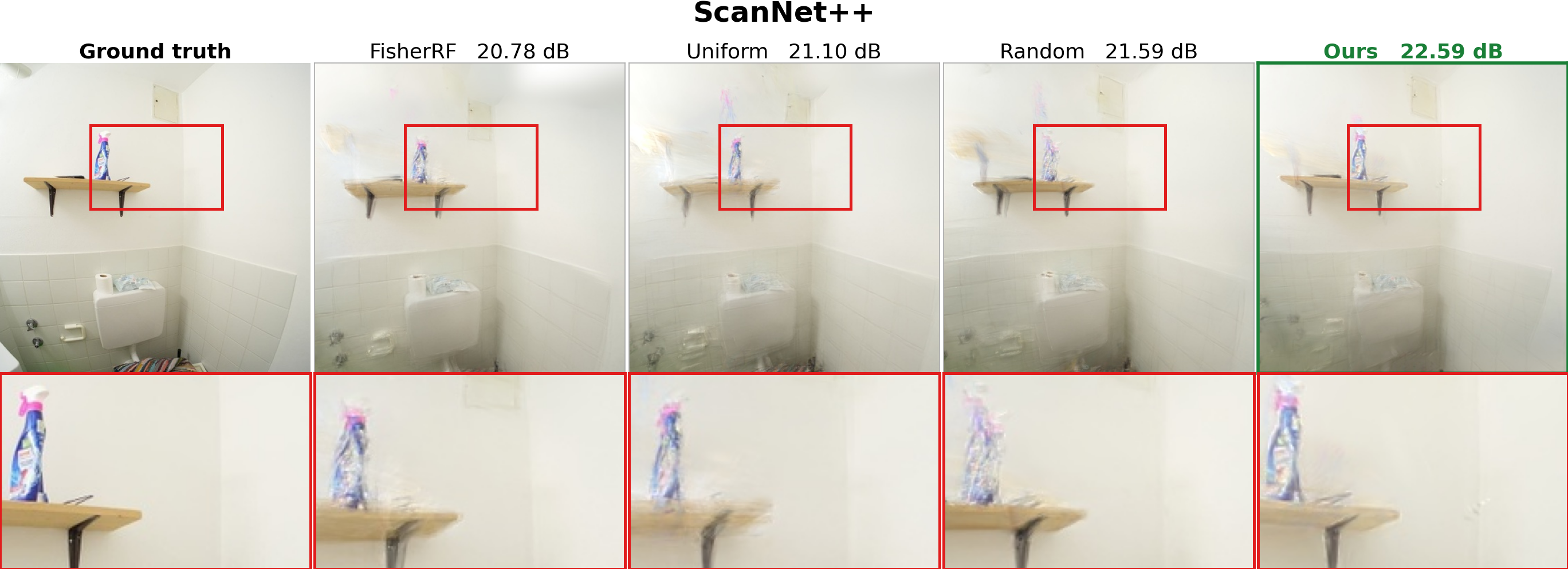}
    \caption{\textbf{Qualitative comparison on ScanNet++} at $K{=}40$.}
    \label{fig:qual_scannetpp}
\end{figure*}

\begin{figure*}[t]
    \centering
    \includegraphics[width=\textwidth]{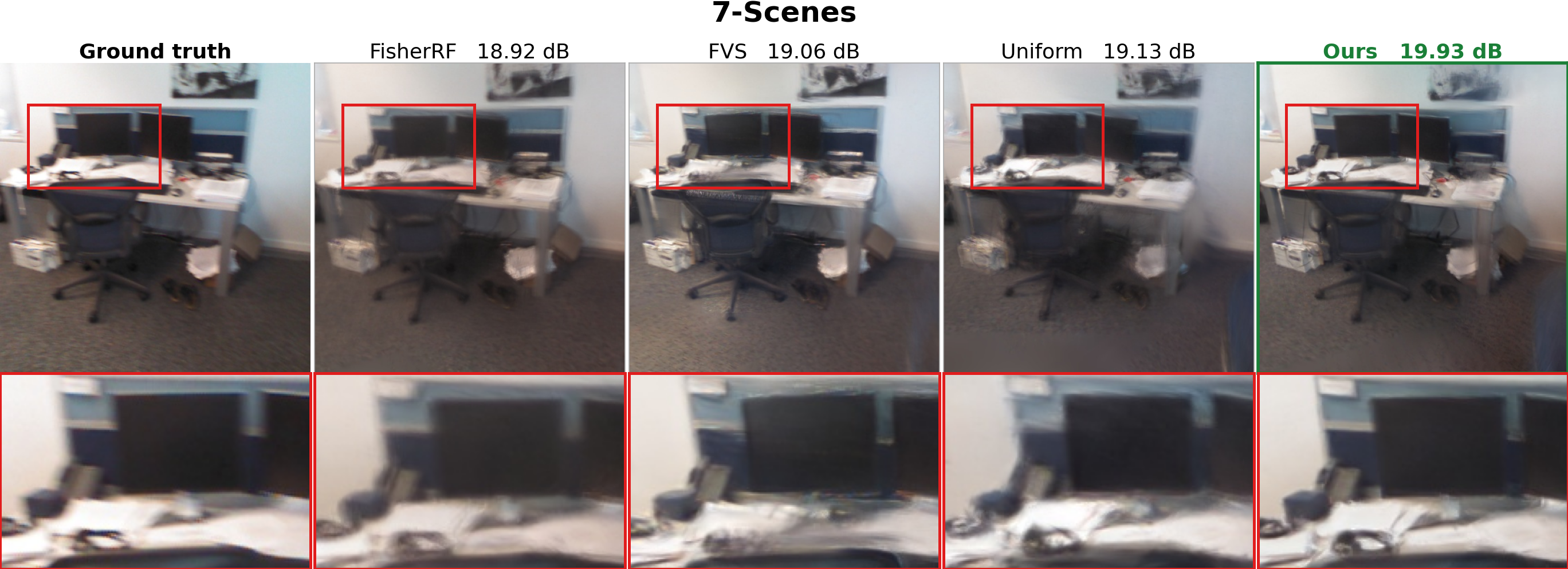}
    \caption{\textbf{Qualitative comparison on 7-Scenes} at $K{=}40$.}
    \label{fig:qual_sevenscenes}
\end{figure*}

\begin{figure*}[t]
    \centering
    \includegraphics[width=\textwidth]{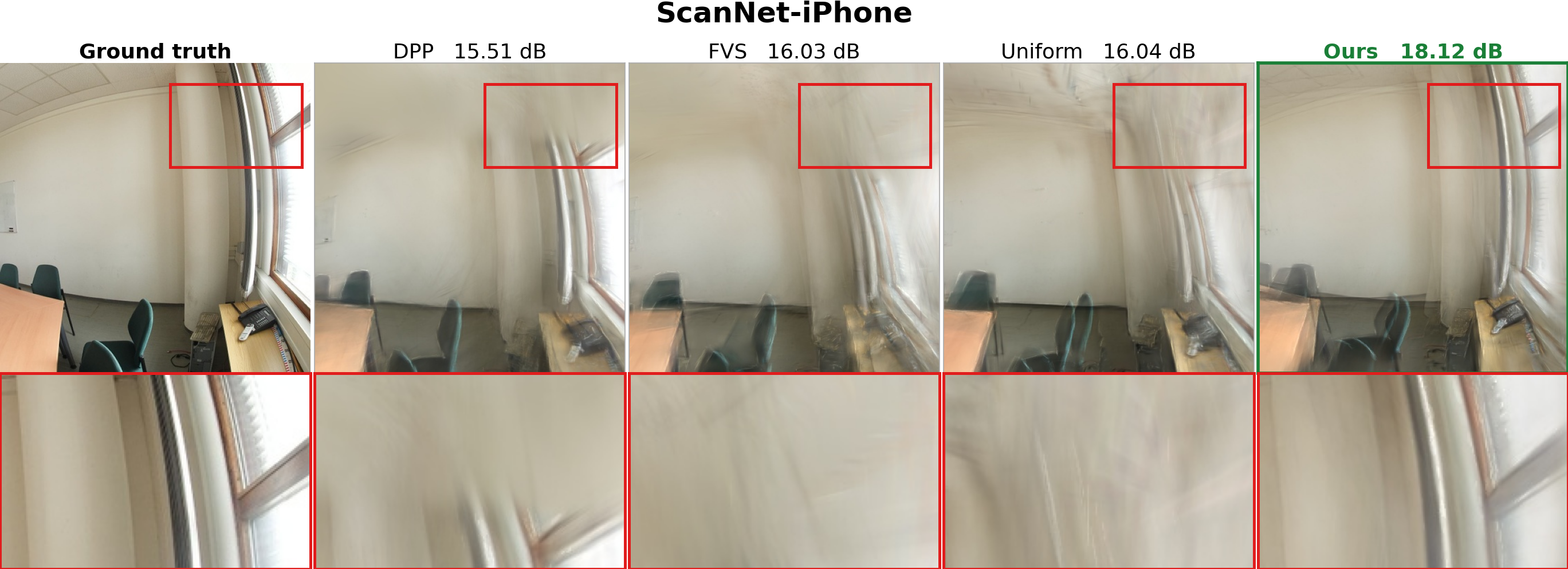}
    \caption{\textbf{Qualitative comparison on ScanNet-iPhone} at $K{=}40$.}
    \label{fig:qual_scanneti}
\end{figure*}

\begin{figure*}[t]
    \centering
    \includegraphics[width=\textwidth]{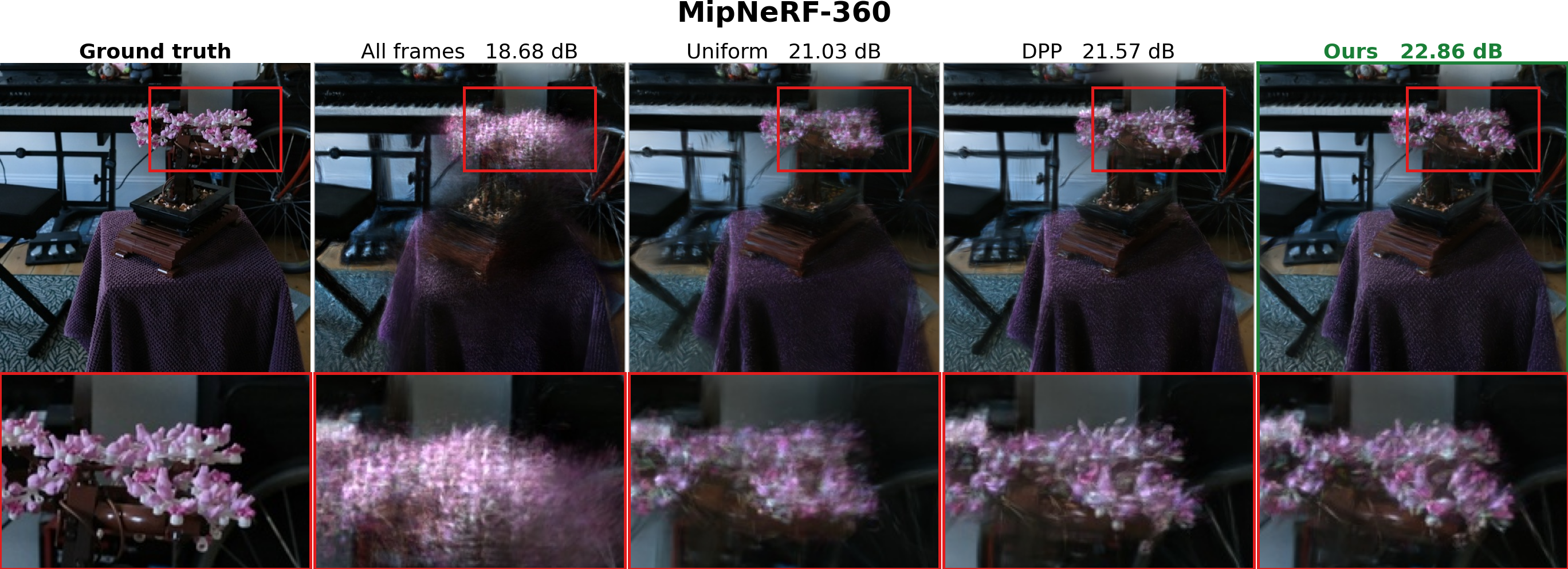}
    \caption{\textbf{Qualitative comparison on Mip-NeRF~360} at $K{=}40$.}
    \label{fig:qual_mipnerf360}
\end{figure*}

We provide additional qualitative comparisons at $K{=}40$ across Tanks and Temples, ScanNet++, 7-Scenes, ScanNet-iPhone, and Mip-NeRF~360, shown in ~\autoref{fig:qual_tandt}, ~\autoref{fig:qual_scannetpp}, ~\autoref{fig:qual_sevenscenes}, ~\autoref{fig:qual_scanneti}, and ~\autoref{fig:qual_mipnerf360}, respectively. These examples complement the quantitative evaluation by illustrating the reconstruction quality obtained from the subsets selected by the different methods across diverse indoor and outdoor scenes.

\end{document}